\documentclass{article} % For LaTeX2e
\usepackage[final]{colm2026_conference}

\usepackage{microtype}
\usepackage{hyperref}
\usepackage{url}
\usepackage{booktabs}
\usepackage{graphicx}

\usepackage{lineno}

\definecolor{darkblue}{rgb}{0, 0, 0.5}
\hypersetup{colorlinks=true, citecolor=darkblue, linkcolor=darkblue, urlcolor=darkblue}

\usepackage{xspace}
\usepackage{listings}

\usepackage{pifont}
\usepackage{booktabs}
\usepackage{tikz}
\usepackage{amsmath}
\usepackage{tabularx}
\usepackage{array}
\usepackage{caption}
\usepackage{wrapfig}
\usepackage{makecell}
\usepackage{tcolorbox}
\usepackage[x11names]{xcolor}

\lstdefinestyle{logstyle}{
  basicstyle=\ttfamily\small,
  columns=fullflexible,
  breaklines=true
}
\lstdefinestyle{promptstyle}{
  basicstyle=\ttfamily\small,
  columns=fullflexible,
  breaklines=true,
  frame=single,
  backgroundcolor=\color{gray!10}
}

\definecolor{darkblue}{rgb}{0, 0, 0.5}
\hypersetup{colorlinks=true, citecolor=darkblue, linkcolor=darkblue, urlcolor=darkblue}

\newcommand{\automat}{\textsc{AutoMat}\xspace}

\newcommand{\internet}{\raisebox{-0.15em}{\includegraphics[height=1em]{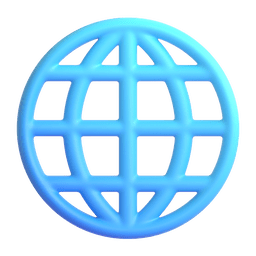}}}
\newcommand{\github}{\raisebox{-0.15em}{\includegraphics[height=1em]{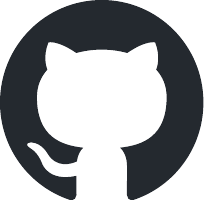}}}
\newcommand{\huggingface}{\raisebox{-0.15em}{\includegraphics[height=1em]{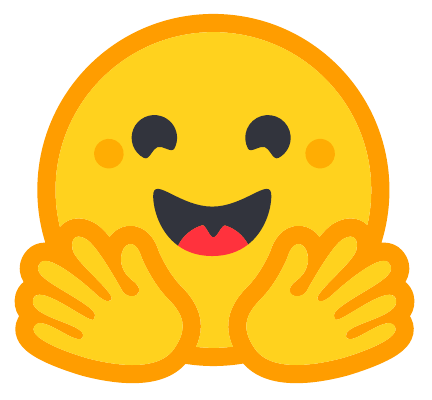}}}

\title{
Can Coding Agents Reproduce \\ Findings in Computational Materials Science?
}

\author{Ziyang Huang, \; Yi Cao\thanks{Equal contribution. $^\heartsuit$Equal advising.}, \; Ali K. Shargh\footnotemark[\value{footnote}], \; Jing Luo,  Ruidong Mei \\
\textbf{Mohd Zaki, Zhan Liu, Wyatt Bunstine, William Jurayj, Somdatta Goswami} \\
\textbf{Tyrel McQueen, Michael Shields, Jaafar El-Awady, Paulette Clancy} \\
\textbf{Benjamin Van Durme, Nicholas Andrews$^\heartsuit$, William Walden$^\heartsuit$, Daniel Khashabi$^\heartsuit$} \\ \\
Johns Hopkins University \\
\texttt{\{zhuang86,noa,wwalden1,danielk\}@jhu.edu} \\
}

\usepackage[table]{xcolor}

\begin{document}

\ifcolmsubmission
\linenumbers
\fi

\maketitle

\begin{center}
  \begin{tabular}{cll}
    \internet & \textbf{Homepage} & \href{https://jhu-clsp.github.io/AutoMat}{\path{https://jhu-clsp.github.io/AutoMat}} \\
    \github & \textbf{Source Code} & \href{https://github.com/JHU-CLSP/AutoMat}{\path{https://github.com/JHU-CLSP/AutoMat}} \\
    \huggingface & \textbf{HF Dataset} & \href{https://hf.co/datasets/jhu-clsp/AutoMat}{\path{https://hf.co/datasets/jhu-clsp/AutoMat}} \\
  \end{tabular}
\end{center}
\vspace{0.5em}

%%%%%%%%%%%%%%%%%%%%%%%%%%%%%%%%%%%%%%%%%%%%%%%%%%%%%%%%%%%%
\begin{abstract}

Large language models are increasingly deployed as autonomous coding agents and have achieved remarkably strong performance on software engineering benchmarks. However, it is unclear whether such success transfers to computational scientific workflows, where tasks require not only strong coding ability, but also the ability to navigate complex, domain-specific procedures and to interpret results in the context of scientific claims. To address this question, we present \automat, a benchmark for evaluating LLM-based agents' ability to reproduce claims from computational materials science. \automat poses three interrelated challenges: recovering underspecified computational procedures, navigating specialized toolchains, and determining whether the resulting evidence supports a claim. By working closely with subject matter experts, we curate a set of claims from real materials science papers to test whether coding agents can recover and execute the end-to-end workflow needed to support (or undermine) such claims. We then evaluate multiple representative coding agent settings across several foundation models. Our results show that current LLM-based agents obtain low overall success rates on \automat, with the best-performing setting achieving a success rate of only 53\%. Error analysis further reveals that agents perform worst when workflows must be reconstructed from paper text alone and that they fail primarily due to incomplete procedures, methodological deviations, and execution fragility. Taken together, these findings position \automat as both a benchmark for computational scientific reproducibility and a tool for diagnosing the current limitations of agentic systems in AI-for-science settings.\footnote{Large language models were used as part of the evaluation pipeline in the form of an LLM-based judge, as described in \S\ref{sec:agent-eval}. Aside from this methodological use, the authors used LLM-based tools only for editorial and implementation assistance. The authors assume full responsibility for all content.}

\end{abstract}

%%%%%%%%%%%%%%%%%%%%%%%%%%%%%%%%%%%%%%%%%%%%%%%%%%%%%%%%%%%%
\section{Introduction}

\begin{figure}[ht]
  \centering
  \includegraphics[width=\linewidth,height=0.32\textheight,keepaspectratio]{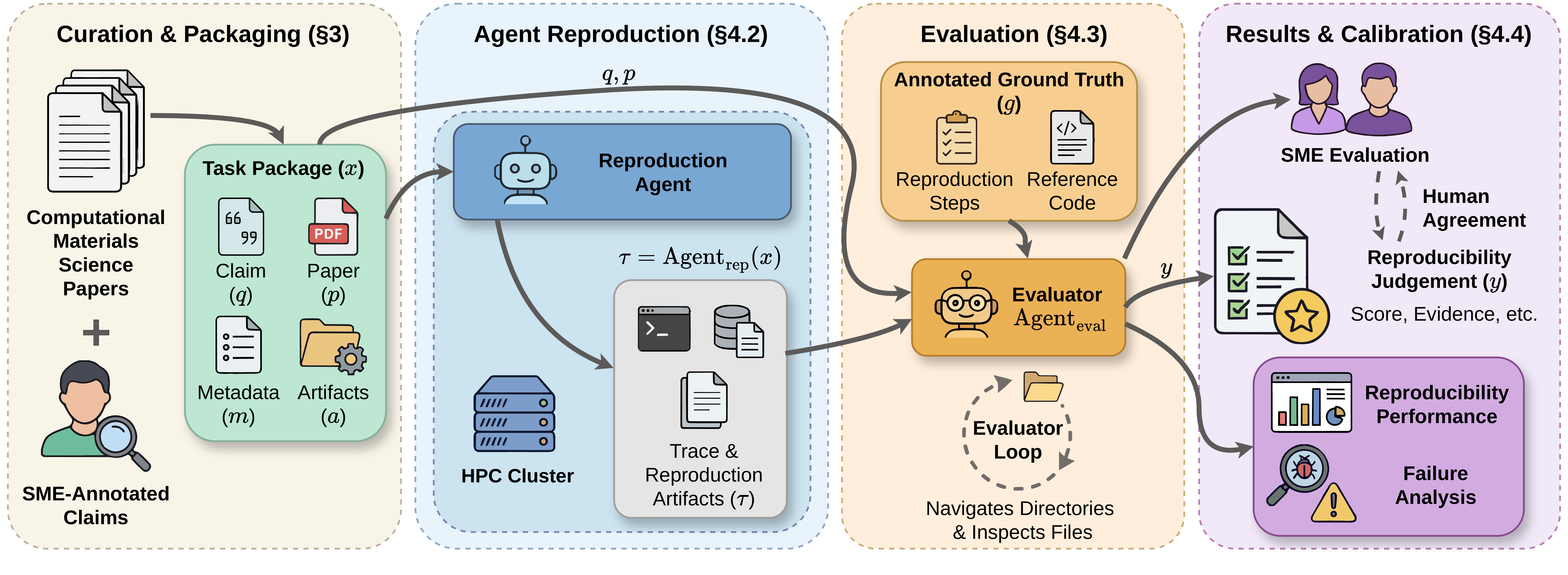}
  \caption{Overview of \automat. Claims from domain experts are packaged into runnable tasks and executed by reproduction agents in an HPC environment. A separate evaluator agent then inspects the resulting trace and artifacts to assign a reproducibility judgment.
  }
  \label{fig:automat-overview}
\end{figure}

Recent advances in large language models (LLMs) have driven rapid progress in code generation and agentic coding assistants, leading to a wave of agents that can write, run, and debug code in interactive environments~\citep{chen2021evaluating, yang2024swe}. These systems have attracted substantial attention because they perform well on software engineering tasks, where requirements are explicit and where correctness is validated via tests, static checks~\citep{jimenez2024swebench}, as well as other intermediate feedback (e.g., linting and compilation). An increasingly mature ecosystem of benchmarks reflects this focus, evaluating how well LLMs generate, modify, and execute code in realistic development settings~\citep{jimenez2024swebench, shah2024stackeval, jain2025livecodebench}.

However, strong performance on software engineering tasks does not necessarily imply strong performance in \textit{reproducing scientific findings}~\citep{siegel2024core, xiang2025scireplicatebench, hu2025repro}. Reproducing a scientific claim requires much more than merely implementing a well-specified function; agents must recover an end-to-end workflow from partial information, navigate domain-specific toolchains, and produce evidence that supports (or undermines) a paper-level result~\citep{krafczyk2021learning}. This makes scientific reproduction a distinct regime---underspecified, tool-rich, and failure-prone---where reasonable-looking outputs can still be scientifically wrong~\citep{antunes2024reproducibility, hassan2025characterising}.

Computational materials science is a particularly stringent domain for this evaluation, since typical claims rely on complex multi-stage pipelines that combine simulation, feature extraction, model training, and post-processing, and often involve specialized external dependencies and tacit methodological choices~\citep{persaud2024reproducibility, li2024reproducibility}. Even when code is released, reproducing a main result may require matching implicit choices that are rarely documented yet often essential in practice~\citep{lejaeghere2016reproducibility}. This makes computational materials science a natural stress test for whether coding agents can function as reliable scientific assistants rather than merely as capable programmers.

In this work, we introduce \automat, a benchmark that operationalizes this challenge for LLM-based agents. \automat comprises 85 carefully selected and annotated computational materials science claims, curated by subject matter experts (SMEs) from recent publications across diverse subdomains. Each instance is framed as an unsupervised reproduction task (\autoref{fig:automat-overview}): given a claim paired with released artifacts (when available), an agent should identify the required computational procedure, implement or adapt the necessary code, configure the environment, execute the workflow, and produce evidence supporting or disputing the claim. To preserve realism while maintaining reliable evaluation, we combine a standardized task format, manual curation, and artifact-grounded assessment, so that the benchmark assesses reproduction \emph{per se} rather than idiosyncrasies of task construction.

We then use \automat to evaluate widely used coding agents in this scientific setting. We compare several off-the-shelf agentic systems as well as a customized, orchestrated workflow across multiple foundation models and configurations. In summary:
\begin{itemize}
    \item We introduce \automat, a benchmark of 85 SME-curated computational materials science claims for evaluating end-to-end scientific reproducibility with LLM agents.
    \item We formulate claim-level scientific reproduction as a runnable benchmark task, with standardized task packaging and artifact-grounded, SME-calibrated assessment.
    \item We present an empirical study of five representative coding agent settings, showing low overall reproduction success, a large gap between from-paper and from-artifact tasks, and recurrent failures in agents' procedure recovery and workflow execution.
\end{itemize}

%%%%%%%%%%%%%%%%%%%%%%%%%%%%%%%%%%%%%%%%%%%%%%%%%%%%%%%%%%%%
\section{Related work}
\label{sec:related_work}
\paragraph{Reproducibility benchmarks} 

Recent benchmarks have begun to evaluate LLM agents on computational reproducibility, though under different task definitions and domains. CORE-Bench~\citep{siegel2024core} evaluates whether agents can regenerate predefined study results from provided code, data, and task materials, while REPRO-Bench~\citep{hu2025repro} focuses on \emph{end-to-end} reproducibility in social science---where agents receive a paper, a reproduction package, and major findings, and judge whether reproduced results support those findings. SciReplicate-Bench~\citep{xiang2025scireplicatebench} instead studies algorithmic reproduction from NLP papers, asking agents to generate code from paper descriptions and evaluating them with execution-based tests and code quality metrics. Similarly, PaperBench~\citep{starace2025paperbench}  targets replication of AI research papers but requires agents to start from scratch; Paper2Code~\citep{seo2026paper2code} aims for full code repository generation from machine learning papers; and ScienceAgentBench~\citep{chen2025scienceagentbench} evaluates agents on individual data-driven discovery tasks drawn from peer-reviewed publications.

\automat is inspired by these efforts but does not merely transpose them onto a new domain: rather than regenerating predefined results from provided materials, producing a faithful implementation of a paper, or solving isolated scientific-coding tasks, \automat targets \emph{fully unsupervised reproduction} of computational materials science \emph{claims}, in which agents must recover or adapt underspecified domain-specific workflows, execute specialized toolchains, and judge whether the resulting evidence supports the claim, possessing a distinct unit of evaluation, domain, and success criterion. \autoref{tab:automat-comparison} summarizes differences between \automat and key prior works.

\begin{table}[ht]
\centering
% \footnotesize
\small
% \fontsize{9pt}{9pt}\selectfont
\setlength{\tabcolsep}{3.5pt}
\renewcommand{\arraystretch}{1.3}
\begin{tabularx}{\columnwidth}{@{}l>{\raggedright\arraybackslash}p{0.30\columnwidth}>{\raggedright\arraybackslash}p{0.18\columnwidth}>{\raggedright\arraybackslash}p{0.12\columnwidth}>{\raggedright\arraybackslash}X@{}}
\toprule
\textbf{Benchmark} & \textbf{Main task} & \textbf{Agent input} & \textbf{Task unit} & \textbf{Domain} \\
\midrule
\makecell[tl]{\textcolor{Coral3}{\textbf{\automat}} \\
(our work)
}
& \textbf{Reproduce a scientific claim:} \textbf{reconstruct and execute} the workflow needed to produce evidence for a specific paper claim.
& Claim, paper, metadata, optional artifacts
& Claim-level
& Computational materials science \\
\arrayrulecolor{gray}\hline
\makecell[tl]{\textbf{REPRO-Bench} \\ \citep{hu2025repro}}
&  Assess whether results reproduced from the paper’s \textbf{reproduction package} are consistent with \textbf{selected major findings}.
& Paper, reproduction package, major findings
& Paper-level, via selected findings/results
& Social science \\
\arrayrulecolor{gray}\hline
\makecell[tl]{\textbf{CORE-Bench} \\ \citep{siegel2024core}}
& Use \textbf{provided code and data} to \textbf{regenerate} specified results.
& Code, data, task materials
& Task-level
& Computer science, social science, medicine \\
\bottomrule
\end{tabularx}
\caption{Comparison between \automat and related reproducibility benchmarks. We distinguish benchmarks by the object of evaluation: \automat targets claim-level scientific reproduction, REPRO-Bench targets paper-level major findings, and CORE-Bench targets predefined study results from provided materials.
}
\label{tab:automat-comparison}
\end{table}

\paragraph{Computational materials science benchmarks for LLMs}
A growing body of work evaluates LLMs in materials science, but most of this work focuses on basic knowledge, question answering, feasibility prediction, or tool use in isolation rather than full, multi-step reproduction of workflows. Matter-of-Fact~\citep{jansen2025matter}, for example, frames materials science progress as a feasibility assessment problem, while MatTools~\citep{liu2025mattools} and SciCode~\citep{tian2024scicode} evaluate materials-related tool use and scientific coding. \automat is complementary to these benchmarks: instead of asking whether a claim is feasible or whether a model can use a tool correctly, \automat asks whether an agent can reconstruct, execute, and validate a computational materials science claim end-to-end.

\paragraph{Coding-agent interfaces}
\automat evaluates a class of systems that are increasingly deployed as general-purpose coding copilots. Terminal-based agents such as Claude Code,\footnote{Claude Code (Anthropic): \url{https://code.claude.com/docs/en/overview}} Codex CLI,\footnote{Codex CLI (OpenAI): \url{https://github.com/openai/codex}} OpenHands CLI,\footnote{OpenHands CLI (OpenHands): \url{https://github.com/OpenHands/OpenHands-CLI}} and Kimi Code\footnote{Kimi Code (MoonshotAI): \url{https://github.com/MoonshotAI/kimi-cli}} support file editing, shell execution, and iterative debugging in local workspaces and are typically positioned as software engineering assistants that work independently on a target task. We treat these systems as representative interfaces and ask whether their strength in software engineering transfers to scientific reproduction, where success depends on workflow recovery, long-horizon execution, and grounded validation---rather than simply passing test cases.

%%%%%%%%%%%%%%%%%%%%%%%%%%%%%%%%%%%%%%%%%%%%%%%%%%%%%%%%%%%%
\section{\automat benchmark}
\label{sec:benchmark}

Here, we describe how \automat is constructed, how claims are packaged into runnable tasks, and what types of reproducibility problems the benchmark contains.

\subsection{Overview}

\begin{figure}[t]
  \centering
  \includegraphics[width=0.88\linewidth]{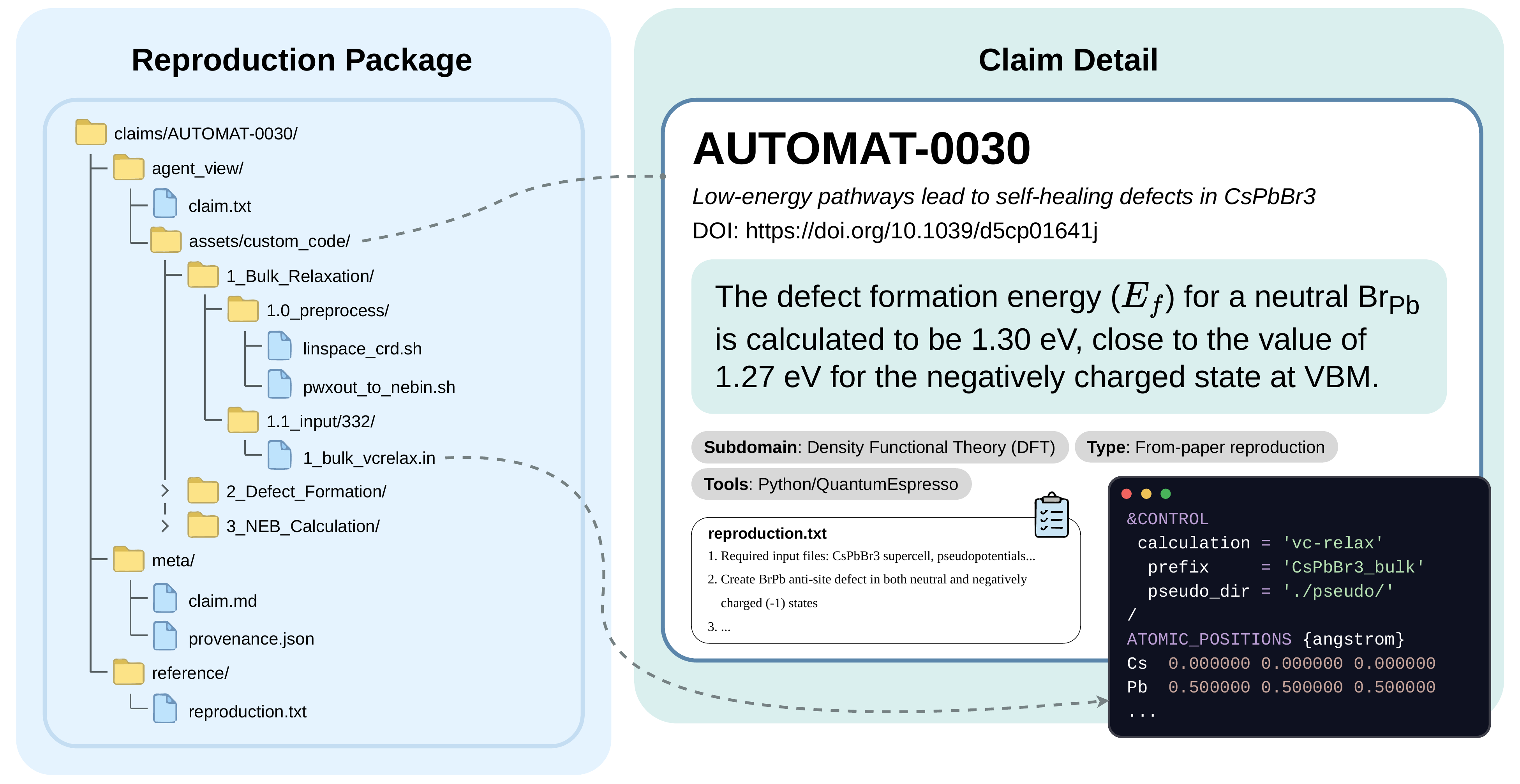}
  \caption{An example claim reproduction package from \automat.
  }
  \label{fig:automat-example}
\end{figure}

\autoref{fig:automat-overview} summarizes \automat's construction. We begin with a curated set of computational materials science papers and SME-annotated claims. Each claim is packaged into a runnable task with associated metadata and available artifacts, executed in a resource-controlled environment representative of modern HPC workflows, and then evaluated with a separate artifact-grounded LLM judge.

\subsection{Benchmark curation}

\autoref{fig:automat-example} shows an example task package. \automat claims span three \emph{reproduction types}, which differ in the amount of procedural recovery required from the agent, and five \emph{verification types}, which differ in how reproducibility is assessed. We cover both in \S\ref{sec:claim-types}.

% \textit{from-paper reproduction}, \textit{from-artifact reproduction}, and \textit{from-artifact interpretation}, each of which introduces unique challenges. We discuss these claim types in detail in \S\ref{sec:claim-types}.

\paragraph{Claim collection and annotation}

\automat tasks are not arbitrary claims extracted from papers. Starting from a collection of 10 computational materials science papers, each task is grounded in first-hand reproduction information obtained from the original authors of each paper, including available artifacts and the procedural guidance needed to reproduce or assess the selected result. Subject matter experts (SMEs, whose backgrounds are detailed in \autoref{tab:sme-background}, \autoref{app:sme-opinion}) then standardize this information into claim-level benchmark instances by identifying the target claim, available inputs, intended reproduction steps, and verification criteria, along with supporting information and expected outcome. Candidate claims are selected to support the source paper's main findings and to be computationally (rather than physically) reproducible. These claims span several major computational materials science subdomains, including Statistical/Machine Learning methods, Density Functional Theory, Molecular Dynamics, and Discrete Dislocation Dynamics,  each of which introduces distinct reproducibility challenges. \autoref{app:sme-opinion} includes short, SME-authored notes that offer further context on these domains and their reproducibility challenges.

We represent each \automat example $x$ as $x = (q, p, m, a)$, where $q$ is the claim text, $p$ is the source paper, $m$ is the metadata file, and $a$ denotes any accompanying artifacts such as scripts, custom codebases, data, or other relevant materials. SMEs are instructed to draft claims centered on numerical results that serve as critical evidence for the paper's main conclusions. This emphasis distinguishes \automat from benchmarks that focus on broad qualitative statements available in natural language critiques \citep{ou-etal-2025-claimcheck}: our goal is to evaluate whether agents can recover and validate \emph{quantitatively checkable} scientific claims.

\begin{wrapfigure}[17]{r}{0.59\linewidth}
  \centering
  \vspace{-0.4cm}
  \includegraphics[width=\linewidth]{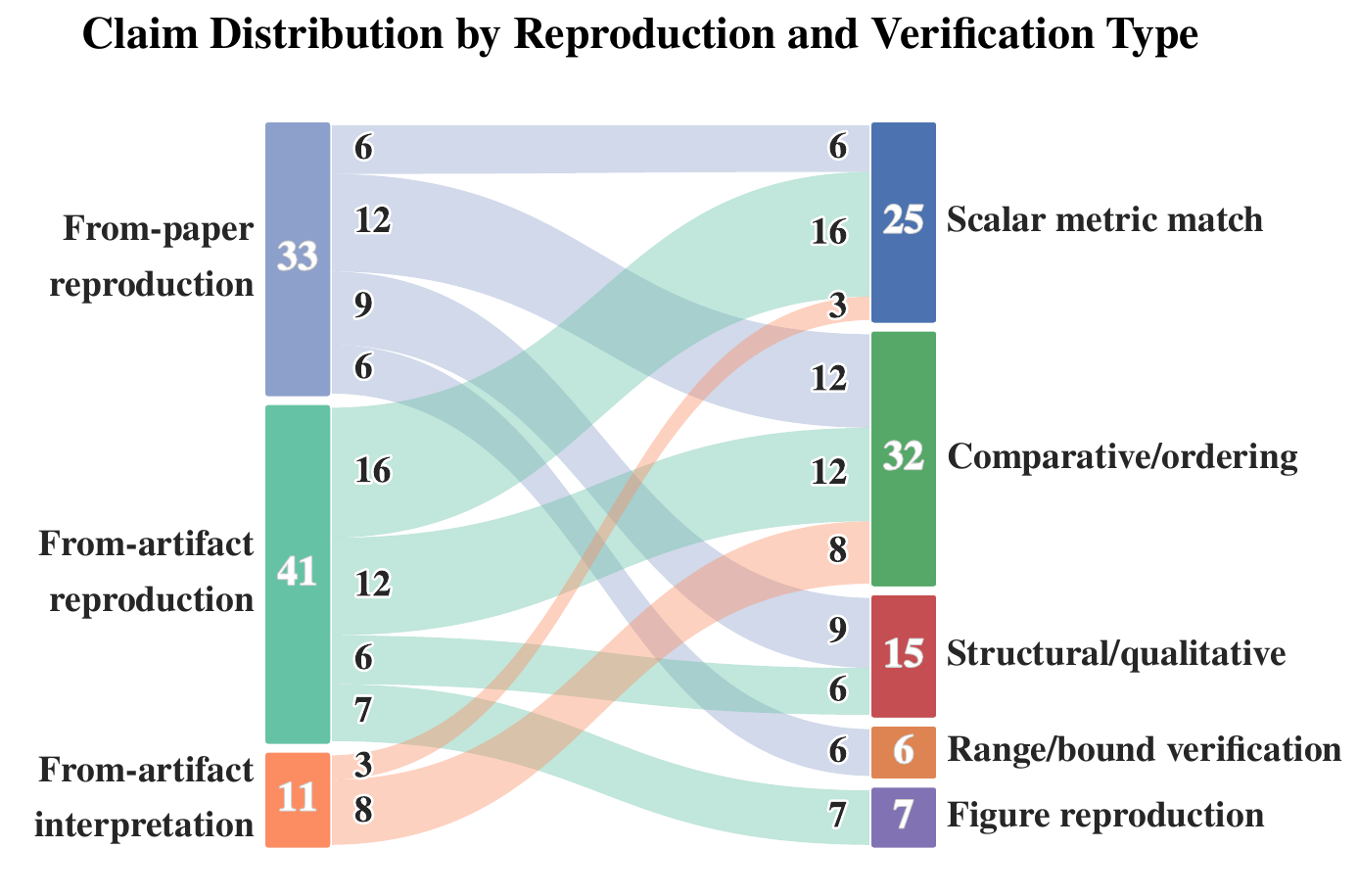}
  \caption{\automat claim distribution by reproduction type (left) and verification type (right).}
  \label{fig:claim-donut}
\end{wrapfigure}

\paragraph{Quality control and solvability}
In total, 96 candidate entries were annotated, and 11 were removed during a secondary quality-control review, yielding the 85 final claims. \autoref{app:qc-solvability} details this review process and discusses how task solvability and agent failures on \automat should be interpreted.

\paragraph{Task packaging}
After collection, each curated claim is converted into a self-contained task package that exposes the claim, paper,\footnote{While we do not observe agents searching for implementations or accessing release artifacts through links in the paper, we remove URLs that might lead to leakage before launching the agents.} metadata file, and any provided artifacts, while concealing provenance and expert annotations for later evaluation. We treat artifacts as optional because reproducibility requirements vary across claims. Some tasks can be reproduced using standard software and configuration inputs described in the source paper, whereas others depend on bespoke codebases, packaged datasets, or specialized scripts. For every claim, a SME-authored, step-by-step reproduction procedure is reserved for evaluation and remains hidden from the agent.

\subsection{Reproduction and verification types}
\label{sec:claim-types}

\automat contains 85 claims spanning three \emph{reproduction types}. \emph{From-paper reproduction} includes claims that are reproducible with generic tools and public resources described in their source paper, or paired with artifacts, such as templates or examples. \emph{From-artifact reproduction} includes claims paired with artifacts consisting of custom codebases, pretrained models, or datasets that are essential to the reproduction path. Finally, \emph{from-artifact interpretation} includes claims with final outputs from simulation workflows, for which reproduction consists primarily of post-hoc analysis and interpretation.

We further categorize each claim based on the \emph{type of verification} required: \emph{scalar metric match}, \emph{comparative/ordering}, \emph{structural/qualitative}, \emph{range/bound verification}, \emph{figure reproduction}. Full taxonomy definitions are provided in Appendix~\ref{app:claim-taxonomy}. \autoref{fig:claim-donut} shows the joint distribution of claims across reproduction and verification types.

%%%%%%%%%%%%%%%%%%%%%%%%%%%%%%%%%%%%%%%%%%%%%%%%%%%%%%%%%%%%
\section{Experiment and evaluation setup}

Below, we detail the five agent settings that we evaluate, the protocol used to produce a reproduction attempt for each claim, and the artifact-grounded evaluation procedure used to score the reproduction runs. We also describe how the LLM-based evaluator is calibrated against SME judgments.

\subsection{Agent settings}

We evaluate five agent settings on \automat. One is an \automat task-specific orchestrated agent (\textbf{Orch.}), built to follow a structured reproduction workflow using the Claude Agent SDK with Claude Sonnet 4.6~\citep{anthropic2026sonnet46}. The other four are general-purpose coding agents: Claude Code (\textbf{CC}) with (1) Claude Opus 4.6~\citep{anthropic2026opus46}, (2) Claude Sonnet 4.6, and (3) Kimi K2.5 (Kimi; \citealp{team2026kimi}); and (4) OpenAI's Codex CLI (\textbf{Codex}) with GPT-5.4~\citep{openai2026gpt54}.

This setup serves two purposes. First, it provides a cross-system view of current end-to-end reproducibility performance. Second, it allows for a comparison between the benchmark-specific orchestrated agent and widely used off-the-shelf coding agents (Claude Code, Codex), enabling us to study the effect of workflow design more directly. Because our experimental budget is limited, these settings should be interpreted as representative end-to-end systems rather than as an exhaustive ablation study across models and interfaces. Full model identifiers, interface versions, and implementation details are provided in \autoref{app:models-toolkits}. Runtime environment and hardware configuration are detailed in \autoref{app:server-spec}.

\subsection{Reproduction attempt}
Given a task instance $x$, consisting of a claim $q$, paper $p$, metadata $m$, and optional artifacts $a$, \automat launches an LLM-based agent to investigate whether the target claim can be reproduced. The agent reads the paper and metadata, plans an investigation, executes commands, inspects outputs, and writes a final report. The run takes place in a resource-controlled execution setting representative of modern HPC workflows, such as environments managed by Slurm~\citep{yoo2003slurm} and (when appropriate) containerized software stacks such as Singularity~\citep{kurtzer2017singularity}. We denote the resulting trajectory as $\tau = \mathrm{Agent}_{\mathrm{rep}}(x)$, where $\tau$ includes the full execution trace, generated files, terminal logs, and the agent's final analysis. This setup captures agents' autonomous investigations end-to-end, rather than their text-only reasoning.

\paragraph{Task-specific orchestrated agent} Among the five evaluated settings, one is a benchmark-specific orchestrated agent (\textbf{Orch.}) designed to impose more structure on the reproduction process. It separates each run into four phases: read-only planning, environment preparation, deterministic execution with LLM-based failure diagnosis, and result extraction with self-assessment. The goal of this design is not to introduce additional scientific knowledge, but to improve auditability and process control compared with the other, more free-form CLI-based agents. We use this system primarily to support the controlled comparison in \S\ref{sec:rq2}. Implementation details are provided in Appendix~\ref{app:models-toolkits}.

\subsection{Artifact collection and evaluation}
\label{sec:agent-eval}
After execution, \automat aggregates the complete evidence produced during the run, including logs, commands, intermediate outputs, reproduction artifacts, and the agent's written report. These materials form the basis for reproducibility assessment. By preserving the full trajectory rather than only the final verdict, \automat enables investigation both of \emph{whether} the agent succeeded and of \emph{how} it carried out its attempt.

A separate LLM-based evaluator agent then judges the reproduction attempt. Unlike a fixed-prompt judge that only consumes a static prompt, the evaluator is allowed to navigate the produced artifact directory, inspect files, and use tools to gather the information needed for assessment. The evaluator receives the SME-annotated ground-truth reproduction steps, $g$, including reference code (when available), along with the original claim $q$, the source paper $p$, and the execution trace $\tau$ with produced artifacts. Formally, we write $y = \mathrm{Agent}_{\mathrm{eval}}(g, q, p, \tau)$, where $y$ is a structured evaluation containing an overall score, dimension-level assessments, and justifications.

The evaluator assigns an overall score on a five-point scale: 5 (\emph{fully reproduced}), 4 (\emph{mostly reproduced}), 3 (\emph{partially reproduced}), 2 (\emph{mostly failed}), 1 (\emph{failed}). We define \emph{successful} reproductions as those that receive an overall score of at least 4. The overall score is supported by evidence along five dimensions: methodological fidelity, execution competence, result accuracy, completeness, and scientific rigor. For each dimension, the evaluator similarly provides a 5-point Likert-scale score and justification, and it may introduce claim-specific sub-dimensions to capture domain-specific criteria. The evaluator also summarizes what the agent did correctly and highlights key errors that prevented a stronger reproduction, explicitly comparing its assessment to the agent's self-judgment when available. All evaluation outputs are written to the artifact directory.

\subsection{Judge calibration against human SMEs}
\label{sec:judge-calibration}

To assess whether automated evaluation is sufficiently aligned with expert judgment for benchmark-scale use, we calibrate the evaluator against blinded assessment from qualified SMEs on a sampled subset ($n=40$) of \automat runs. We construct this subset using stratified sampling to cover different agent systems and reproduction types. For each selected run, SMEs are given the same task materials, execution traces, and produced artifacts used for evaluation. To reduce anchoring effects, SMEs do not see the agent's final answer, self-evaluation, or the LLM judge's assessment. Instead, they score each run independently using the same five-point scale and the same five evaluation dimensions. On this calibration subset, the evaluator achieves a quadratic-weighted kappa~\citep{cohen1968weighted} of 0.69, suggesting substantial agreement, and a within-1 difference accuracy of 0.80---both indicating that the evaluator's judgments are sufficiently aligned with SME assessments to support large-scale analysis. \autoref{app:evaluator-reliability} further reports per-reproduction-type agreement, a confusion matrix, and a multi-label failure-mode audit; the evaluator is slightly conservative relative to SMEs, and we treat failure-mode labels as aggregate diagnostics rather than per-run ground truth.

%%%%%%%%%%%%%%%%%%%%%%%%%%%%%%%%%%%%%%%%%%%%%%%%%%%%%%%%%%%%

\section{Findings and analysis}
We study LLM agents for scientific reproducibility along three axes: overall capability, the role of agent design, and the nature of their failures. Concretely, we aim to answer the following research questions.

\begin{figure}[ht]
  \centering
  \includegraphics[width=\linewidth]{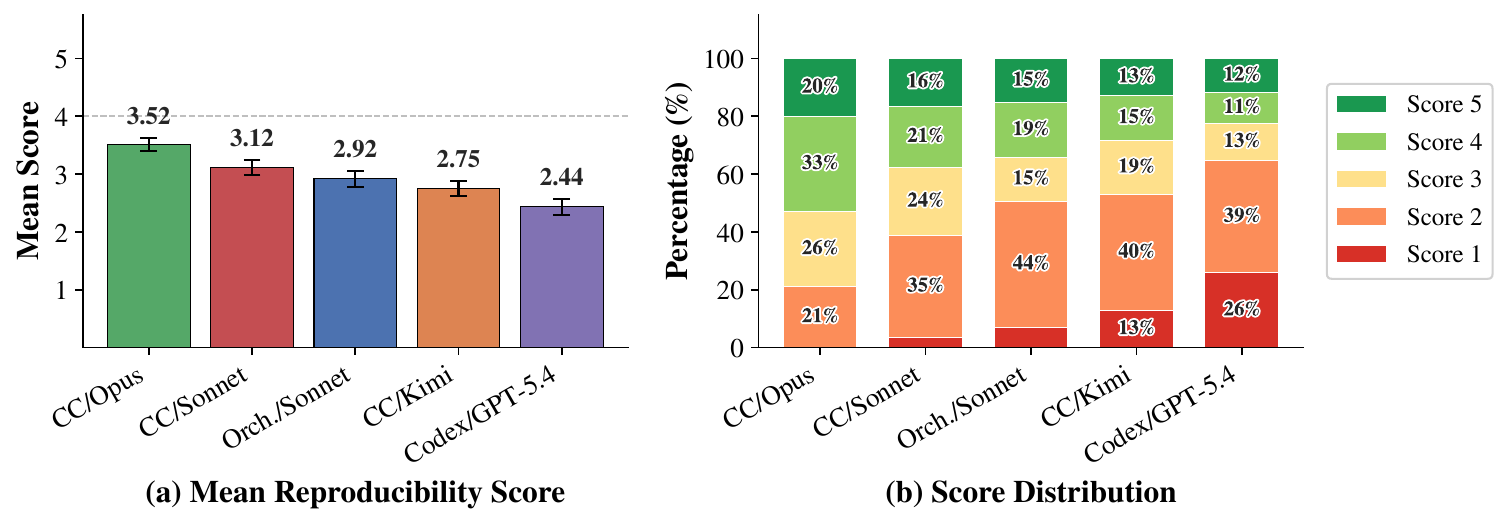}
  \caption{Overall reproducibility performance on \automat across five agent settings. (a) Mean overall reproducibility score assigned by the evaluator on a 1--5 scale. (b) Distribution of evaluator-assigned overall scores across all benchmark instances.
  }
  \label{fig:rq1-overall}
\end{figure}

\begin{itemize}
    \item \textbf{RQ1:} how well can current LLM-based coding agents reproduce computational materials science claims end-to-end?
    \item \textbf{RQ2:} does task-specific orchestration improve reproduction performance over general-purpose coding agents?
    \item \textbf{RQ3:} what failure modes tend to prevent successful reproductions on \automat?
\end{itemize}

\subsection{RQ1: How well can current LLM-based coding agents reproduce computational materials science claims end-to-end?}
\label{sec:rq1}

\textit{Current LLM agents struggle on \automat.} \autoref{fig:rq1-overall}(a) shows that performance remains limited across all systems. Claude Code with Opus is the strongest configuration we evaluate, with a mean overall reproducibility score of 3.52 and a success rate of 53\%, while Codex with GPT-5.4 is the weakest overall, achieving a mean score of 2.44 and a success rate of 23\%. Since scores of 4 and 5 correspond to strong reproductions, this means that \emph{none} of the tested systems is yet a reliable autonomous assistant for scientific reproducibility in this setting, and even the best-performing system succeeds on only about half of all instances. The score distributions in \autoref{fig:rq1-overall}(b) reinforce this picture: across systems, a large fraction of runs receive only mid-range scores, indicating that current agents often make partial progress without reaching a complete and convincing reproduction.

\begin{figure}[t]
  \centering
  \includegraphics[width=\linewidth]{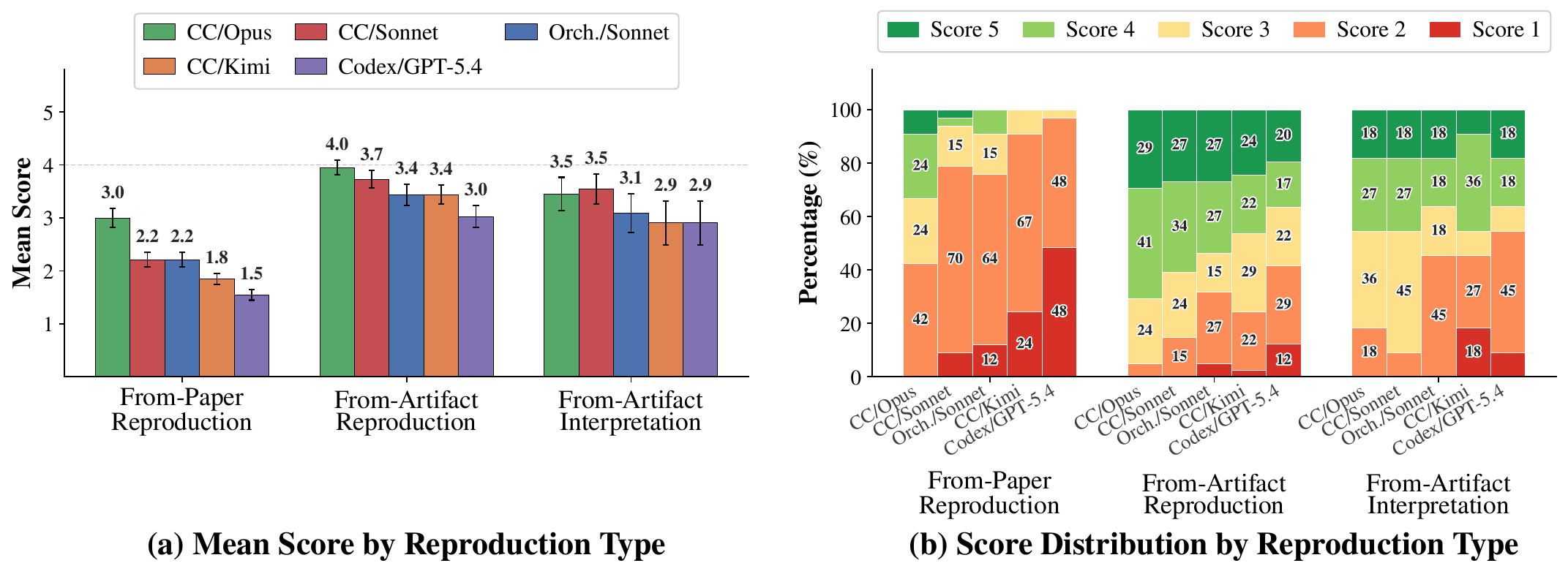}
  \caption{(a) Mean overall reproducibility score for each agent setting, broken down by reproduction type: \emph{from-paper reproduction}, \emph{from-artifact reproduction}, and \emph{from-artifact interpretation}. Error bars denote standard errors. Higher scores indicate better performance. (b) Score distribution within each reproduction type.
  }
  \label{fig:rq1-bytype}
\end{figure}

\textit{Performance varies sharply by reproduction type.} \autoref{fig:rq1-bytype} shows detailed score distributions by reproduction type. Both the mean score and success rate confirm that from-paper reproduction is by far the hardest setting, with mean scores ranging from 1.5 to 3.0 and \emph{near-zero} success rates across all systems except for CC/Opus. This suggests that recovering an underspecified workflow from paper text alone remains a major bottleneck. By contrast, from-artifact reproduction is substantially easier, with mean scores between 3.0 and 4.0 and success rates between 37\% and 71\%, indicating that access to executable artifacts removes much of the ambiguity in the task and shifts the challenges from full workflow recovery to execution and adaptation.

\textit{The from-artifact interpretation setting reveals a complementary limitation.} Even when outputs are already available, performance remains uneven, with success rates ranging from 36\% to 45\% across the systems. This indicates that scientific reproducibility is not only an execution problem but also an interpretation problem: agents should determine whether existing outputs actually support the target claim and justify that judgment convincingly, which remains difficult despite artifacts being available.

To decompose these sources of difficulty, we additionally evaluate a paper-and-artifacts, \emph{no-execution} diagnostic baseline (\autoref{app:no-exec-baseline}). This baseline system typically understands the target claim and expected result, yet frequently fails to recover execution-critical parameters and implementation details, suggesting that \automat's difficulty lies mainly in recovering and executing concrete workflows rather than in understanding claim content.

\subsection{RQ2: Does task-specific orchestration improve performance over general-purpose coding agents?}
\label{sec:rq2} \textit{Task-specific orchestration does not improve overall reproducibility in a controlled comparison.} \autoref{fig:rq2-sonnet} compares our orchestrated agent against Claude Code, both running on Claude Sonnet 4.6. The paired per-claim win/tie/loss breakdown is dominated by ties (44.7\%), with wins and losses remaining relatively balanced. This indicates that the orchestrated workflow does not yield a clear overall advantage in end-to-end success.

\begin{center}
  \begin{minipage}[t]{0.52\linewidth}
    \centering
    \includegraphics[width=\linewidth]{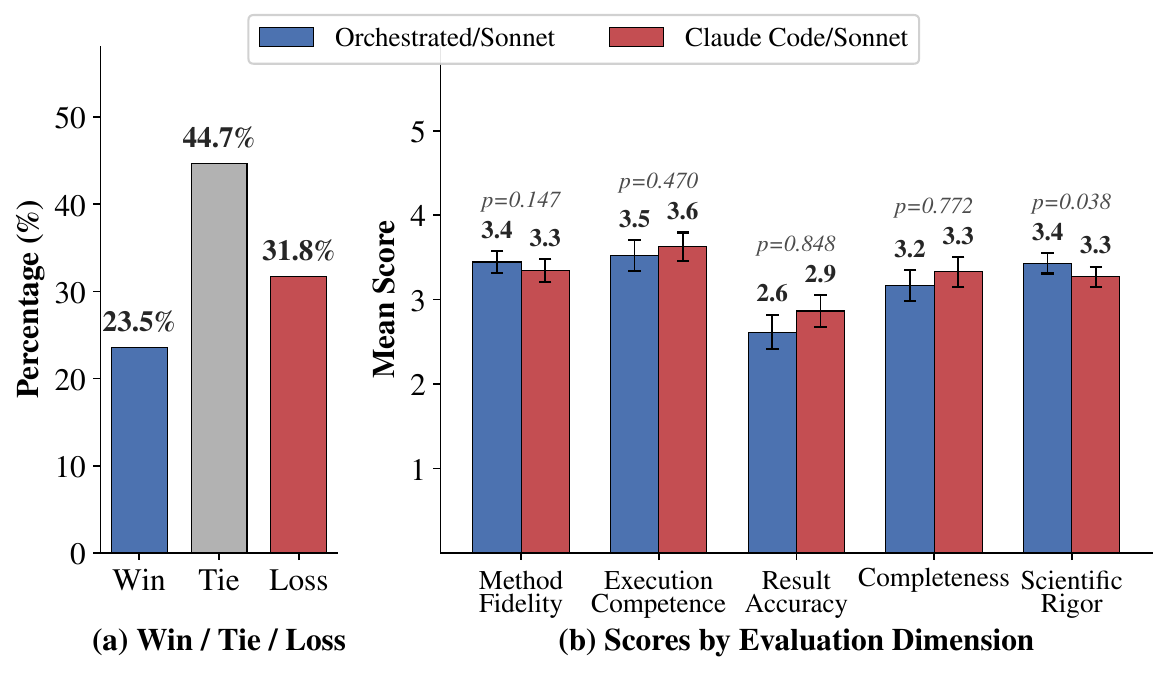}
    \captionof{figure}{Controlled (same-backbone) comparison between the orchestrated agent and Claude Code with Sonnet 4.6. (a) Per-claim win/tie/loss breakdown based on the evaluator's overall reproducibility score. (b) Mean scores by evaluation dimension, with one-sided Wilcoxon signed-rank test $p$-values shown for each dimension.}
    \label{fig:rq2-sonnet}
  \end{minipage}
  \hfill
  \begin{minipage}[t]{0.46\linewidth}
    \centering
    \includegraphics[width=\linewidth]{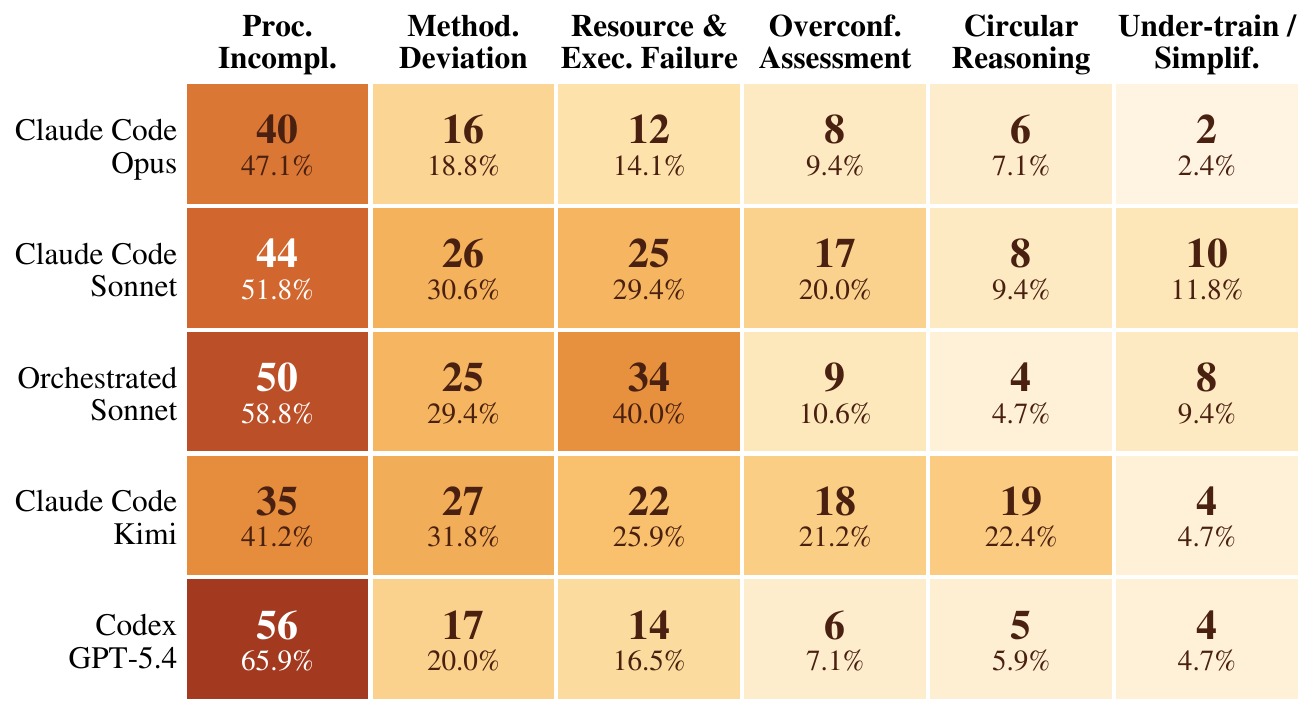}
    \captionof{figure}{Failure-mode heatmap across agent settings on \automat. Each cell shows the count and percentage of all runs in that setting that are assigned the corresponding failure label. A single run may receive multiple failure labels. See Appendix~\ref{app:taxonomy} for full definitions of the failure taxonomy.}
    \label{fig:rq3-failure}
  \end{minipage}
\end{center}

At the dimension level, the orchestrated agent shows directional evidence of higher scientific rigor under the pre-specified one-sided Wilcoxon signed-rank test ($p=0.038$), but this does not reach conventional significance under the corresponding two-sided test ($p=0.076$), and no other dimension is significant under either test (full per-dimension results are reported in \autoref{tab:wilcoxon-dimensions}, \autoref{app:additional-stats}). Overall, the paired comparison suggests that task-specific orchestration changes the \emph{character} of agent behavior rather than producing a robust improvement in final reproduction success. Meanwhile, the orchestrated agent \emph{underperforms} Claude Code with Sonnet on several other evaluation dimensions, suggesting that task-specific orchestration introduces trade-offs---favoring more cautious and evidence-grounded analysis---rather than uniformly improving performance. Since Claude Code supports a more fluid interaction loop than the orchestrated agent, the latter is at a disadvantage in tasks that depend on opportunistic repair and incremental adaptation over long-horizon workflows. By contrast, the orchestrated design appears to favor caution and evidence-grounded analysis, which is consistent with its higher scientific rigor scores.

\subsection{RQ3: What failure modes tend to prevent successful reproductions on \automat?}

\textit{Failures on \automat are dominated by breakdowns in carrying out the intended scientific procedure, rather than by simple downstream mismatches in outputs,} as shown in \autoref{fig:rq3-failure}. Across systems, and among all runs in each setting, the most common failure modes are \emph{procedural incompleteness} and \emph{methodological deviation}, indicating that agents often either fail to complete the necessary workflow or pursue a reproduction path that diverges from the methodology that the claim demands. \emph{Resource and execution failures} are also common, especially for orchestrated agents, showing that long-horizon scientific workflows remain fragile in practice. By contrast, \emph{circular reasoning}, \emph{overconfident assessment}, and \emph{under-training / simplification} occur less frequently overall. Because evaluator--SME agreement on fine-grained failure-mode labels is only moderate, and lower for some settings (\autoref{app:evaluator-reliability}), we interpret these distributions as aggregate trends rather than precise per-run attributions.

The heatmap also reveals distinct behaviors across systems. For example, failures of Codex with GPT-5.4 are particularly driven by procedural incompleteness, suggesting that many failures arise from incomplete workflow execution rather than an inability to reason about the task. The orchestrated agent exhibits fewer instances of overconfidence than its same-backbone counterpart, but still suffers from incomplete procedures and execution breakdowns, consistent with its higher scientific-rigor scores but limited overall success. These patterns broadly help explain the results in \S\ref{sec:rq1}, where agents often make partial progress but fail to recover a complete reproduction.

To make these failure categories concrete, we include three representative case studies. Appendix~\ref{app:case-study-1} illustrates procedural incompleteness: the agent correctly identifies the scientific gap but fails to translate that diagnosis into a workable fallback reproduction strategy.  Appendix~\ref{app:case-study-2} shows how methodological deviation at the dependency-handling stage can cause a complete task failure. Appendix~\ref{app:case-study-3} documents a subtler failure where the agent reproduces the reported scalar metric almost exactly but does so on a deviated evaluation subset, recovering the number without verifying the underlying scientific conclusion.

%%%%%%%%%%%%%%%%%%%%%%%%%%%%%%%%%%%%%%%%%%%%%%%%%%%%%%%%%%%%
\section{Conclusion}
We present \automat, a benchmark for evaluating whether LLM-based coding agents can reproduce computational materials science claims end-to-end. Across 85 expert-curated claims, our results show that current coding agents fall well short of reliable scientific reproducibility, especially when agents must reconstruct underspecified workflows from the text of papers alone. Error analysis further shows that the dominant bottlenecks are procedural incompleteness, methodological deviation, and execution fragility, rather than failures of code generation---suggesting that scientific reproducibility is a meaningfully different regime from conventional software engineering tasks. Our results also suggest that future progress can come from two closely related directions: stronger underlying models, and agent harnesses with stronger workflow planning, environment and tool management, explicit verification protocols, and domain-aware sanity checks. Toward both directions, \automat can serve as a benchmark for end-to-end reproducibility and as a diagnostic testbed for understanding where current agentic systems still fail.

\paragraph{Limitations}

Despite our best efforts, our study has several limitations. \automat currently covers 85 claims from a single (albeit broad) scientific domain, and the distribution of reproduction and verification types reflects the practical realities of expert data collection rather than exhaustive sampling. In addition, like many reproducibility benchmarks, \automat focuses on claims that have been \emph{identified} as reproducible; it does not directly assess whether agents can recognize genuinely \emph{non}-reproducible claims. Finally, while our evaluator agent is calibrated against human experts, automated judgment should be understood as a scalable approximation to expert assessment rather than a replacement for it. Future work should thus expand coverage across domains and claim types, explicitly incorporate non-reproducible or adversarial cases, and conduct large-scale human evaluations.

\section*{Acknowledgments}

This material is based upon work supported by Defense Advanced Research Projects Agency (DARPA) under Contract No. HR001125C0304 and ONR grant (N0001424-1-2089). Any opinions, findings and conclusions or recommendations expressed in this material are those of the author(s) and do not necessarily reflect the views of DARPA. This work was carried out at the Advanced Research Computing at Hopkins (ARCH) core facility (rockfish.jhu.edu), which is supported by the National Science Foundation (NSF) grant number OAC1920103.

\newpage
\bibliography{colm2026_conference}
\bibliographystyle{colm2026_conference}

\newpage
\appendix
\section{Quality control and solvability}
\label{app:qc-solvability}

Each candidate claim is additionally reviewed for appropriateness, artifact sufficiency, concreteness, and suitability for computational reproducibility assessment; claims flagged as low-quality or ambiguous are discussed with the annotators and either revised or dropped. In total, 96 candidate entries were annotated, and 11 were removed during quality control, yielding the 85 final claims.

We also note that several of the source studies were authored by co-authors of this work; this first-hand authorship is, by design, what supplies the artifacts and procedural guidance on which task construction relies (\S\ref{sec:benchmark}).

This process provides strong evidence that the tasks are concrete, assessable, and technically reproducible, though it does not constitute an independent re-reproduction of every source-paper claim under all possible software and hardware environments. Accordingly, agent failure on \automat should be interpreted as failure under the benchmark's public-input setting---the agent did not recover and execute the intended reproduction path from the paper and available artifacts---and not as direct evidence that the original scientific claim is intrinsically irreproducible. We also note that this selection process may favor claims that are central to a paper's conclusions and computationally assessable from the available information.

\section{Taxonomy}
\label{app:taxonomy}

\subsection{Verification types}
\label{app:claim-taxonomy}
\paragraph{Scalar metric match} A claim is assigned to this type if a verification of a single output number (F1 score, $R^2$, accuracy, RMSE, etc.) within a stated or reasonable tolerance is required for a successful reproduction.

\paragraph{Range/bound verification} A claim is assigned to this type if the output of its reproduced outcome should fall within a stated interval instead of a definitive scalar.

\paragraph{Comparative/ordering} Claims of this type typically assert an ordering, trend, or relative relationship between two or more quantities and are verified by confirming that the stated relationship holds in the reproduced results.

\paragraph{Structural/qualitative} Claims of this type focus on structural, geometric, or mechanistic properties of the output.

\paragraph{Figure reproduction} Claims of this type involve reproducing figures from provided code and data, and/or interpretations drawn from the figure.

\subsection{Common failure modes}
\label{app:failure-taxonomy}

\paragraph{Procedural Incompleteness} Skipping critical steps like ionic relaxation, nudged elastic band (NEB) path optimization, or only checking a subset of data cycles mentioned in the claim.

\paragraph{Methodological Deviation} Using incorrect features, pseudopotentials, data splits, or units (e.g., training on the test set, or using projector augmented-wave (PAW) pseudopotentials instead of ultrasoft pseudopotentials (USPP), two common approximations used to represent electron-ion interactions in DFT calculations).

\paragraph{Resource \& Execution Failure} Simulations failing due to inability to set up environment or call simulation tools, Out of Memory errors (OOMs), MPI/parallelization errors, timing out before convergence, etc.

\paragraph{Overconfident Assessment} Behaviors like declaring "Success=True" in the final report despite massive numerical errors or failed underlying calculations.

\paragraph{Circular Reasoning ("Begging the Question")} "Verifying" a claim by hardcoding values from the paper or tuning parameters specifically to match the target result.

\paragraph{Under-training/Simplification} Behaviors like drastically reducing epochs (e.g., 50 vs.\ 3000) or shrinking neural network architectures to fit time limits, leading to invalid models.

\section{LLM and agent configuration}
\label{app:models-toolkits}

\subsection{Model identifiers}
Table~\ref{tab:model-ids} maps the shorthand names used throughout the paper to the official model IDs and providers used in our runs. We keep other hyperparameters and settings unchanged unless otherwise specified.

\begin{table}[ht]
  \centering
  \small
  \begin{tabular}{lll}
    \toprule
    Model Name  & Model ID & Provider/Access \\
    \midrule
    Claude Sonnet 4.6 & \texttt{claude-sonnet-4-6}           & Anthropic\footnotemark[1] \\
    Claude Opus 4.6   & \texttt{claude-opus-4-6}             & Anthropic\footnotemark[1] \\
    GPT-5.4 (reasoning: high)          & \texttt{gpt-5.4-2026-03-05}          & OpenAI\footnotemark[2] \\
    Kimi K2.5         & \texttt{kimi-k2.5}                   & MoonshotAI\footnotemark[3] \\
    \bottomrule
  \end{tabular}
  \caption{Name-to-ID mapping for models used in our benchmark runs.}
  \label{tab:model-ids}
\end{table}

\footnotetext[1]{Anthropic: \url{https://platform.claude.com/docs/en/about-claude/models/overview}. 1M context enabled.}
\footnotetext[2]{OpenAI: \url{https://developers.openai.com/api/docs/models/gpt-5.4}}
\footnotetext[3]{MoonshotAI: \url{https://platform.moonshot.ai/docs/guide/kimi-k2-5-quickstart}}

\subsection{Developer tools and agent interfaces}
\autoref{tab:agent-names} reports versions of the interfaces used in our benchmark runs.

\begin{table}[ht]
  \centering
  \small
  \begin{tabular}{lll}
    \toprule
    Interface  & Version & Developer \\
    \midrule
    Claude Code                  & \texttt{2.1.78}          & Anthropic\footnotemark[4] \\
    Claude Agent SDK for Python  & \texttt{0.1.48}          & Anthropic\footnotemark[5] \\
    Codex CLI                    & \texttt{0.115.0}         & OpenAI\footnotemark[6] \\
    \bottomrule
  \end{tabular}
  \caption{LLM agent frameworks and interfaces used in our benchmark runs.}
  \label{tab:agent-names}
\end{table}

\footnotetext[4]{Claude Code: \url{https://code.claude.com/docs/en/overview}}
\footnotetext[5]{Claude Agent SDK for Python: \url{https://github.com/anthropics/claude-agent-sdk-python}}
\footnotetext[6]{Codex CLI: \url{https://github.com/openai/codex}}

\subsection{Leakage prevention and sandbox isolation}
To prevent evaluation leakage, \automat separates each benchmark instance into an agent-visible payload and an evaluation-only payload using symbolic links. During a reproduction, the agent-visible payload is copied into a temporary sandbox so the agent operates on an isolated copy of the inputs. Evaluation-only materials, including the SME-provided step-by-step reproduction procedure, remain outside the sandbox and are not accessible to the agent. Outputs produced during execution are written to a persistent artifact directory for later inspection.

\subsection{Design and implementation of orchestrated agent}
\label{sec:orch-agent}

\paragraph{Phase 1: read-only planning} In the Plan phase, the agent has read-only access to the claim text, supplementary files, and paper. It may inspect directories, search within files, and extract paper content using a PDF-to-Markdown tool built on Docling~\citep{Docling}, that exposes text, tables, and figures in a searchable format. The goal of this phase is to form a reproduction plan and dependency structure before any state-changing actions are taken: restricting writes during planning reduces accidental file modification, deletion, and premature execution before the workflow has been organized. This constraint is not meant to prevent later recovery---in subsequent phases, the agent can still revise its plan, diagnose errors, and rerun failed steps.

\paragraph{Phase 2: environment preparation} In the Setup phase, the agent is granted write and edit access to prepare an executable reproduction environment. Typical actions include copying relevant inputs into the artifact directory, creating or adjusting input files to match claim parameters, compiling custom code, and installing required packages.

\paragraph{Phase 3: deterministic orchestration with LLM diagnosis} Execution is performed by a Python orchestrator, rather than the LLM, in order to keep process control deterministic. The orchestrator topologically sorts planned steps by their declared dependencies and runs each step as a subprocess with enforced timeouts and captured logs. For each step, the orchestrator records the exit status, runtime, and relevant log excerpts, and also tracks produced outputs in the artifact directory. On failure, the orchestrator invokes a Diagnose sub-phase. A fresh LLM session receives the failed command, error output, and context from completed steps, and recommends one of four actions: (i) retry unchanged, (ii) retry with a modified command or working directory, (iii) skip the step, or (iv) abort. Retried steps are re-executed up to a fixed attempt limit of 3. All diagnosis decisions and outcomes are persisted as structured artifacts in JSON.

\paragraph{Phase 4: result extraction and self-assessment} After execution completes, a new LLM session performs analysis using the original claim text, a summary of execution outcomes, and read access to artifacts. The agent extracts quantitative and qualitative results from outputs (e.g., energies, band gaps, convergence summaries, plots/tables) and compares them to the claim. It returns a structured analysis report with (i) a binary self-judgment, (ii) extracted results, (iii) supporting evidence from artifacts, and (iv) a claim-by-claim comparison.

\section{Runtime environment and hardware configuration}
\label{app:server-spec}
We ran all experiments on Slurm-managed nodes, each equipped with dual Intel Xeon Platinum 8480+ processors (56 cores per CPU), 384 GB of RAM, and, for experiments in the Stat/ML category, one NVIDIA A100 GPU with 80 GB of memory. Although the machine provides 112 physical CPU cores in total, we capped each run at 96 CPU cores. Unless otherwise noted, all jobs were executed with a maximum wall-clock timeout of 72 hours.

The software stack preinstalled in the environment comprises Quantum ESPRESSO 7.5 and LAMMPS 2024.08.29, both installed via conda-forge. We use Python 3.12 as the default runtime, with Python 3.11 reserved for molecular dynamics claims because of dependency incompatibilities under Python 3.12. Our main Python dependencies include, but are not limited to, PyTorch 2.10.0, NumPy 2.4.3, SciPy 1.17.1, scikit-learn 1.8.0, ASE 3.27.0, and pymatgen 2025.10.7.

\section{Prompts used in \automat}
\label{sec:appendix-prompts}

Below, we include the prompts used to launch LLM agents.

% Small boxed prompt placeholder
\newcommand{\promptbox}[1]{%
  \begin{center}%
  \fbox{\begin{minipage}{0.95\linewidth}%
  \small\ttfamily #1
  \end{minipage}}%
  \end{center}%
}

\subsection{System prompt}
\label{app:appendix-system-prompt}

\promptbox{You are a computational material scientist specialized in various simulations. You will run without any human supervision or monitoring. You should work in the artifact directory and do not modify any files outside the artifact directory. In the supplementary files, you may find customized codebases and data that you need to use to reproduce the claim. If not, it may suggest that the claim does not require any supplementary files. If needed, you can copy files from the supplementary materials to the artifact directory and organize them as you wish. A GPU will be available whenever you need it. It is possible that the customized codebases are incomplete or nonfunctional; you should attempt to fix them before concluding that the claim is not reproducible. You should understand that a longer runtime is expected in computational materials science simulations. Be patient, and there's no need to check the status frequently. There is no need to create summary files for me unless you need them as a reference.}

\subsection{Task prompt}
\label{app:appendix-task-prompt}

\promptbox{Reproduce the given claim using the supplementary files and the artifact directory. After you understand the claim and optional codebases, read the paper to see whether it includes any details that might be helpful. Use CUDA\_VISIBLE\_DEVICES=0 to enable GPU use if the computation can be accelerated by the GPU. For MPI-parallelized simulation codes (e.g. pw.x, ph.x, pp.x, bands.x, vasp), use '--oversubscribe' flag for accelerated simulation for MPI simulations. IMPORTANT: Supplementary codebases may contain input files for multiple system sizes or configurations. Always verify that input files match the specific parameters in the claim. Modify input files as needed to match the claim. You should use the ParsePDF tool to parse the PDF files, then read the converted files. Run a full simulation and do not kill the simulation simply because it might take a long time.
    \textbackslash n \textbackslash n CLAIM: \textbackslash n \{claim\}
    \textbackslash n \textbackslash n SUPPLEMENTARY FILES: \{agent\_view\_dir\}
    \textbackslash n \textbackslash n ARTIFACT DIRECTORY: \{artifact\_dir\}
}

\subsection{Evaluation prompt}
\label{app:appendix-eval-prompt}

\promptbox{
You are an expert evaluator for the AutoMat scientific reproducibility benchmark. Your task is to produce a holistic 1--5 reproducibility score for an AI agent's attempt to reproduce a scientific claim, along with a detailed dimensional breakdown.

5 - Fully Reproduced. Criterion fully satisfied, evidence clear and complete. The exact methodology was used, all parameters match the original claim, and results are within the reported tolerance. No meaningful gaps.

4 - Mostly Reproduced. Substantially satisfied with one or two minor deficiencies. Nearly all parameters match, results are within approximately $2\times$ the reported tolerance. The reproduction would be considered successful by domain experts.

3 - Partially Reproduced. Partially satisfied with meaningful gaps. Approximately correct methodology but some parameters differ, results are in the right ballpark but not within tolerance. A partial reproduction.

2 - Mostly Failed. Minimally satisfied with major gaps. The general approach is right but critical parameters are wrong, results differ substantially from claimed values, or significant parts of the claim are unaddressed.

1 - Failed. Not satisfied, no meaningful progress toward reproduction. Wrong methodology, code does not run, no results produced, or the agent did not meaningfully attempt the task.

Evaluation Dimensions. You must evaluate exactly five fixed dimensions.

Methodological Fidelity (methodological\_fidelity): Correct methodology, algorithms, software, and parameters as specified in the claim.

Execution Competence (execution\_competence): Code runs successfully, no critical errors, expected outputs generated.

Result Accuracy (result\_accuracy): Quantitative results match claimed values within reasonable tolerance.

Completeness (completeness): All aspects of the claim addressed, nothing significant skipped.

Scientific Rigor (scientific\_rigor): Understanding of the science, sound decisions under ambiguity, validation of results.

For each dimension, generate 2--5 claim-specific sub-dimensions that are relevant to the particular claim being evaluated. For example, for a DFT claim you might create sub-dimensions such as correct exchange-correlation functional, appropriate k-point grid, and related methodological checks.

Consistency Rules.

Evidence-based only. Score based strictly on what is documented in the trace and artifacts. Do not infer or assume actions the agent might have taken but did not document.

Independence of dimensions. Score each dimension independently. A high score on Execution Competence should not inflate Methodological Fidelity. Avoid halo effects.

Deterministic grade mapping. Score 5 maps to fully\_reproduced. Score 4 maps to mostly\_reproduced. Score 3 maps to partially\_reproduced. Score 2 maps to mostly\_failed. Score 1 maps to failed.

Agent self-assessment comparison. If the agent's own analysis\_result.json is provided, compare your evaluation with the agent's self-assessment and note agreements and disagreements.

}

\section{Perspectives on subdomain-specific reproducibility challenges}
\label{app:sme-opinion}
Here, subject matter experts (SMEs) offer their perspectives on the reproducibility challenges present in four of the subdomains covered in \automat: Density Functional Theory (DFT), Statistical/Machine Learning Methods (Stat/ML), Molecular Dynamics (MD), and Discrete Dislocation Dynamics (DDD). \autoref{tab:sme-background} summarizes the backgrounds and areas of expertise of the SMEs who curated and reviewed \automat claims.

\begin{table}[ht]
  \centering
  \small
  \begin{tabular}{lll}
    \toprule
    SME & Background & Expertise \\
    \midrule
    A & Ph.D.\ student in Chemical and Biomolecular Engineering & DFT, AI/ML \\
    B & Postdoctoral researcher in Mechanical Engineering & MD, Stat/ML \\
    C & Postdoctoral researcher in Mechanical Engineering & DDD, Stat/ML \\
    D & Ph.D.\ student in Mechanical Engineering & DDD \\
    E & Postdoctoral researcher at Hopkins Extreme Materials Institute (HEMI) & DFT, AI/ML \\
    F & Ph.D.\ student in Civil and Systems Engineering & Stat/ML \\
    G & Ph.D.\ student in Physics and Astronomy & DFT \\
    \bottomrule
  \end{tabular}
  \caption{Backgrounds and areas of expertise of the SMEs involved in \automat curation and review.}
  \label{tab:sme-background}
\end{table}

\subsection{Density Functional Theory (DFT)} 
Density Functional Theory (DFT) is a quantum mechanical method used to investigate the electronic structure of many-body systems, particularly atoms, molecules, and solids. It is widely used in materials science for predicting properties such as formation energies ($\Delta E_f$), electronic band structures, and vibrational spectra~\citep{hohenberg1964inhomogeneous, kohn1965self,sholl2022density}.
DFT is an especially useful subdomain for studying reproducibility in materials science since, given the same input file and computational environment, DFT results can be reproduced \emph{exactly}. The challenge of reproducibility with DFT consists chiefly in the fact that expert knowledge is typically \emph{assumed} in describing experimental methodology: it is not standard to include raw input or output files when reporting DFT calculations, nor is it typical to specify in text all the parameters used in a DFT calculation. Thus, to successfully reproduce an experiment involving DFT, knowledge of standard and physically reasonable parameters is often required. Additional challenges arise from the widespread use of custom pseudopotentials, exchange/correlation functionals, and code bases, some of which may be non-trivial to obtain or use. Different DFT packages may also have different standard parameters, requiring knowledge of both packages in order to reproduce a result from one in the other. 

Beyond simply reproducing the numerical result, the interpretation of results can be complex and very nuanced. Depending on the particular use case, the data and conclusions drawn from DFT can vary widely. For example, a scientist interested in exploratory synthesis may calculate the formation energies of different materials and argue that a particular phase is thermodynamically favorable based on these energy differences; this requires consistent calculations of several materials and extraction of their energies. Alternatively, a spectroscopist may be interested in identifying excitations present in a material's Raman spectra, in which case they would require a precise electronic structure calculation, phonon calculation, and potentially more detailed local crystal field or magnetic calculations. Finally, a scientist interested in quantum magnetism may desire spin spiral supercell calculations to determine the magnetic ground state of a non-collinear magnetic system with spin-orbit coupling in order to confirm observations in a neutron scattering experiment.

All these examples illustrate how DFT may be applied in drastically different ways across different fields and how a scientist---whether human or AI---usually must have sophisticated knowledge of those fields and their standard scientific practices to successfully replicate experiments. 
These difficulties also underscore the importance of reproducibility. In recent years, the prediction of new, interesting materials has outpaced experimental validation, and so ensuring predictions are accurate and sensible is increasingly important. Many materials prediction workflows use DFT formation energy and electronic structure calculations to show that materials are synthesizable and interesting. The reproducibility of these results is important in allocating the limited experimental verification capacity where it will be used best. Additionally, DFT frequently complements experimental results to support conclusions drawn about a system, and while some experimental results and DFT outputs may be straightforward to reproduce, the more important thing to reproduce is the connection between the two to corroborate the conclusions drawn from the data.

\subsection{Statistical/Machine Learning methods (Stat/ML)}
In materials science, statistical and machine learning models are often developed as surrogates for other computational or experimental approaches, and thus introduce an additional layer of complexity for reproducibility. Examples include models that approximate density functional theory for property prediction~\citep{xie2018crystal,park2025prediction}, replace Calculation of Phase Diagrams (CALPHAD) for phase stability estimation~\citep{shargh2025temperature,cao2026machine}, emulate molecular dynamics (MD) simulations through machine learned interatomic potentials~\citep{zhang2026machine,zhang2018deep}, and serve as fast surrogates for finite element or phase-field simulations of mechanical response and microstructure evolution~\citep{suttakul2024role,shargh2023interpretable,alhada2024machine}. In addition, ML models can be trained directly on experimental datasets to predict materials properties from microstructure and processing conditions~\citep{yang2024prediction,luoUncertaintyawareMachineLearning2026}.

The final results of these experiments depend not only on the ML model architecture and hyperparameters, but also on how the training data is generated, which in turn requires careful control of the parameters associated with the computational or experimental tools used to produce the data. More importantly, ML models are inherently stochastic, with outcomes influenced by random initialization, data splits, and optimization dynamics. While this issue can be mitigated if trained model checkpoints are shared, it is often the case that only data and code are provided without the corresponding checkpoints.

While setting random seeds can improve consistency to some extent, differences in hardware and software environments can still prevent identical results. Additionally, failure to report key details---such as feature normalization, cross-validation strategies, and hyperparameter tuning procedures---is common and can significantly affect reported performance.

The interpretation of ML results introduces further challenges. Reported metrics are meaningful only with respect to the exact evaluation protocol used to generate them. As illustrated in Appendix~\ref{app:case-study-3}, a reproduction can closely match a reported scalar metric while failing to verify the intended scientific conclusion because the evaluation subset was constructed differently. This makes it essential to clearly define not only the model, dataset, and parameters associated with data generation, but also the full evaluation protocol, including preprocessing, feature identification, normalization, filtering, and metric computation. As ML models are increasingly used as surrogates to guide materials discovery, ensuring reproducibility is critical for establishing confidence in both the predictions and the underlying scientific conclusions.

\subsection{Molecular Dynamics (MD)}
While DFT is the "first-principles" method of choice for many materials science applications, it is often computationally prohibitive for large systems or long time scales. Molecular Dynamics (MD) simulations are a complementary computational method that balances computational efficiency with physical accuracy by using classical mechanics to model the atomic interactions, providing insights into the dynamic behavior of materials in larger systems and longer time scales, which are widely used to study phenomena such as phase transitions, diffusion, and mechanical deformation~\citep{thompson2022lammps, horstmann2022structural}.

MD presents a distinctive reproducibility challenge in materials science. In principle, MD simulations are deterministic given a well-defined initial configuration, interatomic potential, boundary conditions, and integration scheme. In reality, however, reproducibility is often limited by incomplete reporting of simulation details and sensitivity to modeling choices. Key parameters, such as thermostat and barostat settings, time step, equilibration scheme, and details of the implemented MD packages, are often not fully specified. Interatomic potentials, particularly machine-learned or custom potentials, introduce additional difficulties when training data and fitting procedures are not publicly available. Even when nominally the same potential is used, differences in parameterization can lead to deviations in predicted behavior.

Beyond numerical reproducibility, interpreting MD results also poses challenges. Post-processing analyses often rely on software such as OVITO or VMD, or on custom user-developed scripts, both of which require user-defined thresholds and methodological choices that can significantly affect outcomes if not explicitly reported. As a result, different analysis pipelines applied to the same MD trajectory can yield different conclusions. For example, quantities such as the number of grains, phase fractions, or bond statistics can vary substantially depending on the chosen parameters.

As computational resources continue to expand, MD is increasingly used to guide materials design and inform experiments. This trend thereby amplifies the importance of reproducibility in both the simulation setups and subsequent analyses, which is essential for establishing reliable physical insights and enabling broader scientific adoption.

\subsection{Discrete Dislocation Dynamics (DDD)}
Discrete Dislocation Dynamics (DDD) is a specialized field that focuses on understanding and predicting the mechanical behavior of metals by explicitly modeling dislocations (a type of line defect in metallic materials) as segments and by modeling their evolution. Working in this field requires not only a solid foundation in plasticity, dislocation theory, and computational solid mechanics, but also technical expertise in scientific programming (particularly in C++) and in parallel computing. Given these requirements, the barrier to entry in this field is relatively high compared with many other areas of computational materials science, which has limited the number of researchers working in this area. As a result, the volume of works based on DDD that we have access to is significantly smaller than that in many other areas of computational materials science.

\section{Case studies}
\label{app:case-studies}
Here, we present three detailed and representative case studies from distinct \automat instances. The first two highlight clear agent failure modes in density functional theory workflows, while the third illustrates a subtler interpretive risk in statistical/machine learning evaluation: reproducing a reported scalar metric without faithfully recovering the underlying evaluation protocol. While familiarity with the relevant areas of materials science is important for understanding each study in its full depth, we also provide less technical summaries of the problems for non-materials science readers in blue boxes below.

\subsection{Case study 1: procedural incompleteness}
\label{app:case-study-1}
This appendix analyzes a representative failure from our benchmark, in which the agent correctly identified missing inputs for a density functional theory (DFT) task, but did not proceed with a feasible partial reproduction strategy.

\begin{tcolorbox}[
    colback=blue!5!white,
    colframe=blue!50!black,
    title={\textbf{TL;DR for non-materials science readers}},
    fonttitle=\normalsize,
  ]
  This is an end-to-end scientific workflow task, not a single-script execution task. The key point is that the agent is not only asked to ``run code,'' but to \emph{reconstruct a full simulation-and-analysis pipeline} from paper text and partial artifacts.
  
  \begin{itemize}
      \item \textbf{Input:} a claim, the source paper, and supplementary artifacts, plus the required run configuration recovered from the paper (structures, parameters, and interaction settings such as pseudopotentials).
      \item \textbf{Output:} reproducible evidence that the simulated migration behavior and structural analysis are consistent with the paper's conclusion.
      \item \textbf{End-to-end process:} read and extract requirements from the paper, plan executable steps, collect or reconstruct missing inputs, run and monitor long-horizon simulations, post-process results, and compare evidence against the published claim.
  \end{itemize}
\end{tcolorbox}

\subsubsection{Task description}

The claim (ID: \texttt{AUTOMAT-0037}) originates from a DFT study of halide ion migration in CsPbBr$_3$ perovskites:

\begin{quote}
\textit{``For both the charged and neutral pathways, the coordination polyhedra for the starred points adopt the following structures: monocapped trigonal prism (Coordinate 1), pentagonal bipyramid (Coordinate 10), monocapped trigonal prism (Coordinate 19), pentagonal bipyramid (Coordinate 28), pentagonal bipyramid (Coordinate 37).''}
\end{quote}

This claim concerns Nudged Elastic Band (NEB) calculations for Br interstitial migration in a $2\times2\times2$ CsPbBr$_3$ supercell (161 atoms), and requires identifying local coordination environments at five characteristic points along the migration path.

\subsubsection{Reproduction attempted}
Table~\ref{tab:case-study-scores} summarizes the evaluation across our five assessment dimensions plus the overall score.

\begin{table}[h]
\centering
\caption{Evaluation scores for \texttt{AUTOMAT-0037}}
\label{tab:case-study-scores}
\begin{tabular}{lcc}
\toprule
\textbf{Dimension} & \textbf{Score (1-5)} & \textbf{Assessment} \\
\midrule
Methodological Fidelity & 2 & Correctly identified, not executed \\
Execution Competence & 1 & Only pseudopotential download \\
Result Accuracy & 1 & No results produced \\
Completeness & 1 & 0/5 coordinates verified \\
Scientific Rigor & 3 & Accurate diagnosis, no partial attempt \\
\midrule
\textbf{Overall} & \textbf{1} & \textbf{Failed} \\
\bottomrule
\end{tabular}
\end{table}

The agent's execution (total runtime: 8 minutes 40 seconds) is summarized below.

	\textbf{Legend:} [\ding{51}] completed, [\(\circ\)] partially completed, [\ding{55}] not completed.

\begin{itemize}
    \item [\ding{51}] \textbf{Environment verified}: confirmed Quantum ESPRESSO v7.5 (\texttt{pw.x}, \texttt{neb.x}) availability.
    \item [\ding{51}] \textbf{Claim localization}: identified the target claim in Section 2.2.2 and Fig.~5 of the paper.
    \item [\ding{51}] \textbf{Artifact inspection}: examined all provided supplementary input files.
    \item [\ding{51}] \textbf{Gap diagnosis}: detected mismatch between required interstitial inputs and available vacancy inputs.
    \item [\ding{51}] \textbf{Infrastructure setup}: downloaded GBRV pseudopotentials.
    \item [\(\circ\)] \textbf{Recovery attempt}: started a Python script to add interstitial structures but did not complete a runnable pipeline.
    \item [\ding{55}] \textbf{End-to-end reproduction}: did not produce validated outputs for the five target coordinates.
\end{itemize}

\begin{tcolorbox}[
    colback=blue!5!white,
    colframe=blue!50!black,
    title={\textbf{TL;DR for non-materials science readers}},
    fonttitle=\normalsize,
  ]
  The agent correctly recognizes that the provided files were insufficient to directly run the target simulation. However, instead of trying a reasonable fallback strategy, it determined that the reproduction was impossible. This is akin to recognizing that a recipe is missing several ingredients, but giving up immediately rather than attempting to reconstruct the missing steps from the available instructions. In this case, the main failure was not misunderstanding the scientific goal, but failing to translate the diagnosis into action.
\end{tcolorbox}

\subsubsection{Analysis of agent's behavior}
The agent demonstrates accurate scientific understanding of the problem in several respects. First, the agent correctly identifies that the claim requires Br \textit{interstitial} migration NEB (161 atoms), which is distinct from the \textit{vacancy} migration NEB (159 atoms) provided in the supplementary materials:

\begin{lstlisting}[style=logstyle]
    The NEB input provided (vacancy_migration_neb.in, 159 atoms 
    with +1 charge) is for vacancy migration, NOT for the 
    interstitial migration described in the claim (which would 
    require 161 atoms).
\end{lstlisting}

The agent also identifies that the available relaxation inputs used incompatible supercell dimensions, as summarized in~\autoref{tab:file-mismatch}.

\begin{table}[h]
\centering
\caption{Supplementary file analysis}
\label{tab:file-mismatch}
\begin{tabular}{lccc}
\toprule
\textbf{File} & \textbf{Atoms} & \textbf{Supercell} & \textbf{Compatible} \\
\midrule
\texttt{1\_bulk\_vcrelax.in} & 360 & $3\times3\times2$ & \ding{55} \\
\texttt{2\_ibr\_relax.in} & 361 & $3\times3\times2$ & \ding{55} \\
\texttt{vacancy\_relax.in} & 159 & $2\times2\times2$ & \ding{55} \\
\texttt{vacancy\_migration\_neb.in} & 159 & $2\times2\times2$ & \ding{55} \\
\textit{Required (missing)} & 161 & $2\times2\times2$ & N/A \\
\bottomrule
\end{tabular}
\end{table}

Despite an accurate diagnosis, the agent concluded that the reproduction was infeasible without exhausting all available options. The agent's stated reasons were:

\begin{enumerate}
    \item Missing NEB input files for Br interstitial migration
    \item Absence of the five relaxed interstitial endpoint structures
    \item Prohibitive computational cost of generating inputs from scratch
\end{enumerate}

% \subsection{Root Cause}
% Evidence from the trace indicates that the agent possesses relevant crystallographic knowledge sufficient to reconstruct the missing file, as evidenced by an attempt to add a Br interstitial using a Python script. The attempt, however, was abandoned without completion.

% To verify that the agent possesses but fails to invoke relevant chemistry knowledge, we conducted the following supplementary experiments.

% \subsubsection{Direct}

% \subsection{Summary}

% This case study exemplifies a failure mode we term \textbf{``diagnosis without remediation''}---the agent accurately characterized why reproduction could not proceed with provided files but failed to leverage available knowledge and resources to attempt partial validation. Key findings include:

% \begin{enumerate}
%     \item \textbf{Granularity misalignment}: High-level task failure blocked exploration of feasible sub-tasks
%     \item \textbf{Dormant knowledge}: Chemistry knowledge existed but was not activated in the problem-solving context
%     \item \textbf{Conservative termination}: Complete infeasibility was conflated with any-attempt infeasibility
% \end{enumerate}

% These findings suggest that current agents' limitations in computational materials science may stem less from knowledge gaps than from \textbf{context-sensitive knowledge invocation}---the ability to recognize when and how to apply domain knowledge to overcome practical obstacles.

\subsection{Case study 2: methodological deviation}
\label{app:case-study-2}

In another failure case (AUTOMAT-0028), the agent correctly identified that the workflow was impeded by missing pseudopotential files, but misdiagnosed \emph{why} those files were missing and treated the situation as fatal rather than applying standard expert workarounds. 

\begin{tcolorbox}[
    colback=blue!5!white,
    colframe=blue!50!black,
    title={\textbf{TL;DR for non-materials science readers}},
    fonttitle=\normalsize,
  ]
  The agent was impeded by missing technical files, but the real problem was not that the files were unavailable. Rather, the agent made an incorrect assumption about how those files should be named, continued searching for nonexistent files, and then incorrectly concluded that the task could not continue. In other words, a small early mistake triggered a chain of errors that prevented a successful reproduction.
  \end{tcolorbox}

\subsubsection{Task description}

The claim (ID: \texttt{AUTOMAT-0028}) originates from a study of halide perovskite defects:

\begin{quote}
\textit{``Our neutral calculations resulted in large formation energies of 6.1 eV and 1.3 eV for $Pb_{Br}$ and $Br_{Pb}$ anti-site defects, respectively.''}
\end{quote}

A successful reproduction requires:
\begin{itemize}
    \item Quantum ESPRESSO DFT code (pw.x) with PBE exchange-correlation functional
    \item Ultrasoft pseudopotentials for three elements: \verb|br_pbe_v1.4.uspp.F.UPF|, \verb|cs_pbe_v1.uspp.F.UPF|, \verb|pb_pbe_v1.uspp.F.UPF|
    \item Pristine CsPbBr$_3$ bulk structure: $3\times3\times2$ supercell (360 atoms, Pnma symmetry, lattice constants $a$=8.23~\AA, $b$=8.18~\AA, $c$=11.73~\AA)
    \item Two relaxed anti-site defect structures: PbBr (Pb atom at Br site) and BrPb (Br atom at Pb site)
    \item SCF energy calculations with DFE formula: $E_f = [E_{\text{defect}} - E_{\text{pristine}}] + \mu_{\text{Br}}$ (chemical potential correction)
\end{itemize}

\subsubsection{Reproduction attempted}

Table~\ref{tab:case-study-dfe-scores} summarizes the evaluation across our five assessment dimensions plus the overall score.

\begin{table}[h]
\centering
\caption{Evaluation scores for AUTOMAT-0028}
\label{tab:case-study-dfe-scores}
\begin{tabular}{lcc}
\toprule
\textbf{Dimension} & \textbf{Score (1-5)} & \textbf{Assessment} \\
\midrule
Methodological Fidelity & 2 & QE/PBE/USPP correctly identified \\
Execution Competence & 3 & Diagnosis complete; no computation executed \\
Result Accuracy & 1 & No DFT results produced \\
Completeness & 2 & No calculations \\
Scientific Rigor & 2 & Flawed reasoning \\
\midrule
\textbf{Overall} & \textbf{2} & \textbf{Mostly Failed} \\
\bottomrule
\end{tabular}
\end{table}
%%%%

Upon inspecting the reasoning trace, the agent hallucinated the existence of several pseudopotentials. As discussed, the correct dependency identifiers are

\begin{itemize}
    \item Br: \texttt{br\_pbe\_v1.4.uspp.F.UPF}
    \item Cs: \texttt{cs\_pbe\_v1.uspp.F.UPF}
    \item Pb: \texttt{pb\_pbe\_v1.uspp.F.UPF}
\end{itemize}

Notably, Br uses v1.4, while Cs/Pb use v1, and these exact strings serve as the dependency identifiers for the run. The agent, however, implicitly enforced a false uniformity constraint:

\begin{quote}
“If Br is v1.4, then Cs and Pb should also be v1.4.”
\end{quote}

This caused the agent to keep attempting to search for and download v1.4 files for all three elements, resulting in 404 errors for Cs and Pb because the requested artifacts do not exist. Thus, due to this false versioning assumption, the agent incorrectly reported that the pseudopotentials cannot be found. The failure therefore stemmed less from missing scientific prerequisites than from an early dependency-identification error that propagated into an incorrect infeasibility judgment.

\subsection{Case study 3: evaluation protocol drift}
\label{app:case-study-3}

Unlike the first two case studies, this run (\texttt{AUTOMAT-0007}) was graded as mostly reproduced because it produced a near-exact numerical match to the reported machine learning metric. However, the match was obtained using a deviated evaluation subset, making the case a useful example of how an ML reproduction can recover the right scalar result while failing to independently verify the intended scientific conclusion.

\begin{tcolorbox}[
    colback=blue!5!white,
    colframe=blue!50!black,
    title={\textbf{TL;DR for non-materials science readers}},
    fonttitle=\normalsize,
  ]
  This case is not a simple execution failure. The agent matched the paper's reported accuracy almost perfectly, but it did so on a differently constructed evaluation subset. In other words, it reproduced the right number without fully verifying that the number still meant the same thing scientifically.
  \end{tcolorbox}

\subsubsection{Task description}

The claim (ID: \texttt{AUTOMAT-0007}) originates from a machine learning study of phase prediction in refractory multi-principal element alloys:

\begin{quote}
\textit{``Using an absolute deviation tolerance of 0.07 per phase, the trained MLP achieves an accuracy of approximately 0.89 on the test dataset filtered to $1250 < T < 1450$.''}
\end{quote}

A faithful reproduction required four elements to be recovered together:
\begin{itemize}
    \item The correct trained MLP checkpoint and architecture
    \item The correct prediction metric: a sample counts as correct only if all 8 phase outputs satisfy an absolute error tolerance of \texttt{0.07}
    \item The correct physical filter: evaluate only test samples in the temperature window $1250 < T < 1450$
    \item The resulting filtered accuracy, which should be approximately \texttt{0.89}
\end{itemize}

In the reference workflow, temperature is recovered from the last input feature using the denormalization $T = x \times (1891 - 850) + 850$, so the target interval $1250 < T < 1450$ corresponds to a normalized range of approximately $(0.384, 0.576)$.

\subsubsection{Reproduction attempted}

Table~\ref{tab:case-study-ml-scores} summarizes the evaluation across our five assessment dimensions plus the overall score.

\begin{table}[h]
\centering
\caption{Evaluation scores for \texttt{AUTOMAT-0007}}
\label{tab:case-study-ml-scores}
\begin{tabular}{p{4.1cm}cp{5.3cm}}
\toprule
\textbf{Dimension} & \textbf{Score (1-5)} & \textbf{Assessment} \\
\midrule
Methodological Fidelity & 3 & Correct model and metric, wrong feature/filter protocol \\
Execution Competence & 5 & Clean end-to-end evaluation \\
Result Accuracy & 5 & Near-exact accuracy match \\
Completeness & 4 & All components addressed; filtering only approximate \\
Scientific Rigor & 2 & Flawed subset selection \\
\midrule
\textbf{Overall} & \textbf{4} & \textbf{Mostly Reproduced} \\
\bottomrule
\end{tabular}
\end{table}

The key quantitative outcome is straightforward:
\begin{itemize}
    \item Claimed filtered accuracy: approximately \texttt{0.89}
    \item Reproduced filtered accuracy: \texttt{0.8894}
    \item Absolute difference from claim: \texttt{0.0006}
    \item Reported filtered sample count: \texttt{8,411}
\end{itemize}

\subsubsection{Analysis of agent's behavior}

The agent succeeded on several important parts of the task. It loaded the correct checkpoint, reconstructed the MLP architecture, and implemented the all-or-nothing tolerance metric correctly. These steps establish that the reproduced number did not arise from a trivial implementation error.

The methodological deviation emerged when recovering the evaluation subset. The agent searched across 35 input features for a plausible temperature-like column and selected feature \texttt{18}. It then evaluated several candidate normalized ranges, including $(0.35, 0.60)$, $(0.38, 0.58)$, and $(0.40, 0.55)$, and retained the combination that produced a result closest to the target answer.

Two aspects of this behavior are especially important:
\begin{enumerate}
    \item \textbf{Temperature identification was heuristic rather than reference-based.} The reference workflow uses the last input feature and an explicit denormalization rule, whereas the agent inferred a candidate temperature feature from the data and selected a different column.
    \item \textbf{Range selection was target-conditioned.} Rather than independently deriving the normalized interval associated with $1250 < T < 1450$, the agent effectively chose the feature/range combination that minimized the difference between the computed accuracy and the claimed value.
\end{enumerate}

This distinction matters because the claim is not merely that the model can achieve an accuracy near \texttt{0.89} on some subset of the test set. The claim is that it achieves that performance on a physically meaningful temperature window. Once feature identification and filtering deviate from the reference evaluation protocol, the meaning of the scalar metric changes even if the number itself remains nearly identical.

For that reason, this case supports a weaker conclusion than the paper's intended statement. The run demonstrates that a subset can be constructed on which the model attains an accuracy near \texttt{0.89} under the correct tolerance metric. It does \emph{not} fully verify that this performance holds on the paper's intended $1250 < T < 1450$ regime. More broadly, the case illustrates why preprocessing, feature identification, normalization, filtering, and metric computation are part of the scientific result in ML-based materials workflows rather than peripheral implementation details.

\section{Evaluator reliability analyses}
\label{app:evaluator-reliability}

\subsection{Agreement by reproduction type and score confusion}

By design, each calibration run was scored by the SME who authored the corresponding source study, since reproduction quality often hinges on methodological details that the original authors are best positioned to assess. \autoref{tab:judge-by-type} breaks down evaluator--SME agreement on the 40-run calibration subset (\S\ref{sec:judge-calibration}) by reproduction type. Agreement is highest for from-artifact interpretation tasks and lowest for from-paper reproduction tasks, as expected: the latter involve more underspecified workflows and more ambiguous partial-credit judgments.

\begin{table}[ht]
  \centering
  \small
  \begin{tabular}{lcccccc}
    \toprule
    Reproduction type & $n$ & QWK & Exact & Within-1 & Mean abs.\ diff. & Bias \\
    \midrule
    From-paper reproduction     & 15 & 0.592 & 66.7\% & 73.3\% & 0.600 & $-0.600$ \\
    From-artifact reproduction  & 13 & 0.698 & 76.9\% & 76.9\% & 0.462 & $-0.154$ \\
    From-artifact interpretation & 12 & 0.833 & 91.7\% & 91.7\% & 0.167 & $-0.167$ \\
    \midrule
    Overall                     & 40 & 0.689 & 77.5\% & 80.0\% & 0.425 & $-0.325$ \\
    \bottomrule
  \end{tabular}
  \caption{Evaluator--SME agreement on the 40-run calibration subset, broken down by reproduction type. QWK denotes quadratic-weighted kappa; Exact and Within-1 denote exact and within-1 score agreement; Bias denotes the mean signed difference between evaluator and SME scores (negative values indicate that the evaluator scores lower).}
  \label{tab:judge-by-type}
\end{table}

\begin{table}[ht]
  \centering
  \small
  \begin{tabular}{lcccccc}
    \toprule
     & \multicolumn{5}{c}{Evaluator score} & \\
    \cmidrule(lr){2-6}
    SME score & 1 & 2 & 3 & 4 & 5 & Total \\
    \midrule
    1 & 0 & 0 & 1  & 0 & 0 & 1 \\
    2 & 1 & 7 & 0  & 0 & 0 & 8 \\
    3 & 5 & 0 & 11 & 0 & 0 & 16 \\
    4 & 0 & 1 & 0  & 8 & 0 & 9 \\
    5 & 0 & 0 & 1  & 0 & 5 & 6 \\
    \midrule
    Total & 6 & 8 & 13 & 8 & 5 & 40 \\
    \bottomrule
  \end{tabular}
  \caption{Confusion matrix between SME and evaluator overall reproducibility scores on the 40-run calibration subset.}
  \label{tab:judge-confusion}
\end{table}

\autoref{tab:judge-confusion} reports the full confusion matrix between SME and evaluator overall scores. The evaluator is somewhat conservative relative to human SMEs: its scores are lower on average (overall bias $-0.325$), especially for from-paper reproduction tasks. Follow-up discussions with SMEs suggest that materials scientists often give more credit to partial reproductions that recover the core scientific workflow or make scientifically reasonable progress, because they are familiar with the practical difficulty of reproducing computational materials science results. By contrast, the LLM evaluator tends to apply the rubric more strictly when exact outputs or methodological details are missing. We report this systematic bias explicitly and regard it as a limitation of automated evaluation.

\subsection{Multi-label failure-mode audit}
\label{app:failure-mode-audit}

Beyond overall scores, we audit the evaluator's multi-label failure-mode annotations against SME annotations on 73 stratified-sampled runs. Because each run may receive multiple failure labels, exact-set agreement is overly strict; we therefore report example-level F1 and Jaccard overlap in addition to exact-set and any-overlap agreement. As shown in \autoref{tab:failure-mode-audit}, the evaluator achieves overall example-level F1 of 0.606 and Jaccard overlap of 0.572, indicating that it often identifies overlapping failure modes even when it does not exactly reproduce the full human tag set. Agreement is weaker for some autonomous coding-agent settings, such as Codex with GPT-5.4 and Claude Code with Kimi K2.5. From manual inspection, low-agreement cases often involve multi-causal traces in which several labels are simultaneously plausible: for example, a run that stops after identifying a missing dependency could reasonably be labeled as procedural incompleteness, resource and execution failure, or methodological deviation, depending on whether one emphasizes the missed fallback strategy, the failed execution path, or the dependency-handling decision.

We emphasize that the multi-label failure-mode audit is not the basis for the benchmark's headline results: the overall reproducibility score is calibrated separately against blinded SME judgments (\S\ref{sec:judge-calibration}). We therefore use failure-mode labels to identify aggregate diagnostic trends across settings, rather than as definitive per-run causal attributions.

\begin{table}[ht]
  \centering
  \small
  \begin{tabular}{lccccc}
    \toprule
    Setting & $n$ & Any-overlap & Example F1 & Jaccard & Exact-set \\
    \midrule
    CC/Opus        & 14 & 64.3\% & 0.576 & 0.554 & 50.0\% \\
    CC/Sonnet      & 15 & 80.0\% & 0.656 & 0.589 & 40.0\% \\
    Orch.          & 15 & 86.7\% & 0.822 & 0.800 & 73.3\% \\
    CC/Kimi        & 14 & 57.1\% & 0.488 & 0.452 & 35.7\% \\
    Codex/GPT-5.4  & 15 & 53.3\% & 0.478 & 0.456 & 40.0\% \\
    \midrule
    Overall        & 73 & 68.5\% & 0.606 & 0.572 & 47.9\% \\
    \bottomrule
  \end{tabular}
  \caption{Agreement between evaluator- and SME-assigned multi-label failure modes on 73 stratified-sampled runs. Any-overlap counts a run as agreeing if the SME and evaluator share at least one failure-mode tag, or if both assign no failure-mode tag.}
  \label{tab:failure-mode-audit}
\end{table}

\section{No-execution diagnostic baseline}
\label{app:no-exec-baseline}

To better understand the sources of task difficulty in \automat, we implement a paper-and-artifacts, \emph{no-execution} diagnostic baseline using Claude Opus 4.6. This baseline receives the same claim, paper, metadata, and supplementary artifacts as the reproduction agents, but is restricted to read-only inspection: it can inspect and search files, but cannot execute code, launch simulations, install packages, modify files, or generate reproduction artifacts. We do not view this as a stronger reproduction baseline than the end-to-end coding agents evaluated in the main paper, since it cannot, by construction, produce executable evidence for many \automat tasks; rather, it serves as a diagnostic ablation that isolates paper understanding, artifact inspection, and final result interpretation from execution.

Across benchmark tasks, the no-execution baseline shows strong paper-level understanding but weaker alignment with the executable reproduction path. The LLM evaluator scores its claim comprehension at 4.88/5, suggesting that the baseline usually understands what scientific result is being targeted, and it identifies the expected result in 73/85 cases. However, its ability to recover execution-critical details is weaker: it achieves a ground-truth parameter-fidelity score of 3.83/5 and an artifact-awareness score of 3.94/5, and on 28/85 claims it receives high claim-comprehension scores but low method and parameter scores---that is, it understands the scientific claim but fails to specify the concrete parameters or implementation details required to reproduce it. In manual review, we find that the baseline often misses execution-specific details such as unstated hyperparameters, data preprocessing operations, checkpoint-loading conventions, and the canonical artifact or intermediate file---precisely the kinds of details that often require interacting with the released artifacts through execution rather than reading the paper and static files alone. This suggests that reasoning without execution can make a task appear well understood, and potentially reproducible, while still missing details required for actual reproduction: much of \automat's difficulty lies not in misunderstanding what the paper claims, but in recovering and executing the concrete computational workflow required to produce evidence.

This diagnostic analysis has clear limitations: it uses a single no-execution model, relies on an LLM evaluator that is not separately calibrated against SMEs for this setting, and should not be interpreted as a full reproduction baseline. Its purpose is to decompose task difficulty, not to replace the end-to-end coding-agent evaluation.

\section{Additional statistical analyses}
\label{app:additional-stats}

\autoref{tab:wilcoxon-dimensions} reports per-dimension Wilcoxon signed-rank tests for the controlled comparison between the orchestrated agent and Claude Code with Claude Sonnet 4.6 (\S\ref{sec:rq2}). The one-sided tests correspond to the pre-specified alternative hypothesis that the orchestrated agent performs better, motivated by RQ2's directional question of whether task-specific orchestration improves reproduction performance. The two-sided tests additionally account for the possibility that orchestration degrades performance. The orchestrated agent shows directional evidence of higher scientific rigor under the one-sided test, but no dimension reaches conventional significance under the two-sided test.

\begin{table}[ht]
  \centering
  \small
  \begin{tabular}{lccl}
    \toprule
    Dimension & One-sided $p$ & Two-sided $p$ & Interpretation \\
    \midrule
    Methodological fidelity & 0.147 & 0.294 & Not significant \\
    Execution competence    & 0.470 & 0.940 & Not significant \\
    Result accuracy         & 0.848 & 0.304 & Not significant \\
    Completeness            & 0.772 & 0.456 & Not significant \\
    Scientific rigor        & 0.038 & 0.076 & Directional/suggestive \\
    \bottomrule
  \end{tabular}
  \caption{Per-dimension one-sided and two-sided Wilcoxon signed-rank test $p$-values for the same-backbone comparison between the orchestrated agent and Claude Code with Sonnet 4.6.}
  \label{tab:wilcoxon-dimensions}
\end{table}

\end{document}